\documentclass{article}

\usepackage[preprint,nonatbib]{neurips_2022}
\pdfoutput=1

\usepackage[utf8]{inputenc} 
\usepackage[T1]{fontenc}    
\usepackage{hyperref}       
\hypersetup{
    colorlinks=true,
    linkcolor=blue,
    filecolor=magenta, 
    urlcolor=cyan,
    citecolor=green,
    pdfpagemode=FullScreen,
}


\usepackage{url}            
\usepackage{booktabs}      
\usepackage{amsfonts}       
\usepackage{nicefrac}       
\usepackage{microtype}      
\usepackage{xcolor}         
\usepackage{textcomp}

\usepackage{rotating}
\usepackage{titlesec}
\usepackage{amsmath}
\usepackage{breqn}
\usepackage{bbm}
\usepackage{amsthm}
\usepackage{amssymb}
\usepackage{mathtools}
\usepackage{xspace}
\usepackage{enumitem}
\usepackage[hang,flushmargin]{footmisc}
\usepackage{algorithm}
\usepackage{algpseudocode}
\usepackage{subfigure}
\usepackage{wrapfig}
\usepackage[
hypcap=false,
skip=0pt,
]{caption}
\setlist{leftmargin=*}
\setlist[itemize]{noitemsep}
\setlist[enumerate]{noitemsep}
\newcommand{\TS}{\mathbf{S}}
\newcommand{\methodSplit}{split}
\newcommand{\methodCenteredSuffix}{BC}
\newcommand{\methodBudgetSuffix}{B}
\newcommand{\methodErrorSuffix}{E}
\newcommand{\methodCentered}{TQA-\methodCenteredSuffix\xspace}
\newcommand{\methodBudget}{TQA-\methodBudgetSuffix\xspace}
\newcommand{\methodError}{TQA-\methodErrorSuffix\xspace}
\newcommand{\methodname}{TQA\xspace}
\definecolor{falured}{rgb}{0.7, 0.15, 0.15}

\newcommand{\dataCLAIM}{\texttt{CLAIM}\xspace}
\newcommand{\dataMIMIC}{\texttt{MIMIC}\xspace}
\newcommand{\dataCOVID}{\texttt{COVID}\xspace}
\newcommand{\dataGEFCom}{\texttt{GEFCom}\xspace}
\newcommand{\dataGEFComR}{\texttt{GEFCom-R}\xspace}
\newcommand{\dataEEG}{\texttt{EEG}\xspace}
\newcommand{\baselineCQR}{CQRNN\xspace}
\newcommand{\baselineMADSplit}{LASplit\xspace}
\newcommand{\baselineCFRNN}{CFRNN (Split)\xspace}
\newcommand{\baselineQRNN}{QRNN\xspace}
\newcommand{\baselineDPRNN}{DPRNN\xspace}

\newcommand{\defeq}{\vcentcolon=}
\DeclarePairedDelimiter\ceil{\lceil}{\rceil}
\DeclarePairedDelimiter\floor{\lfloor}{\rfloor}

\newcommand\fontred[1]{{\color{red}#1}}

\newtheorem{theorem}{Theorem}[section]

\newtheorem{definition}{Definition}

\usepackage{xcolor}

\title{Conformal Prediction with Temporal Quantile Adjustments}

\author{%
  Zhen Lin \\
  University of Illinois at Urbana-Champaign\\
  Urbana, IL 61801 \\
  \texttt{zhenlin4@illinois.edu} \\
  \And
  Shubhendu Trivedi\thanks{Primary affiliation where research was initiated and carried out.}\\
  MIT\\
  Cambridge, MA 02139 \\
  \texttt{shubhendu@csail.mit.edu} \\
  \And 
  Jimeng Sun \\
  University of Illinois at Urbana-Champaign\\
  Urbana, IL 61801 \\
  \texttt{jimeng@illinois.edu} \\
}

\begin{document}

\maketitle

\begin{abstract}
We develop Temporal Quantile Adjustment (\methodname), a general method to construct efficient and valid prediction intervals (PIs) for regression on cross-sectional time series data.
Such data is common in many domains, including econometrics and healthcare. 
A canonical example in healthcare is predicting patient outcomes using physiological time-series data, where a population of patients composes a cross-section. 
Reliable PI estimators in this setting must address two distinct notions of coverage:
\textit{cross-sectional} coverage across a cross-sectional slice, and \textit{longitudinal} coverage along the temporal dimension for each time series. 
Recent works have explored adapting Conformal Prediction (CP) to obtain PIs in the time series context. 
However, none handles both notions of coverage simultaneously.
CP methods typically query a pre-specified quantile from the distribution of \textit{nonconformity scores} on a calibration set.
\methodname adjusts the quantile to query in CP at each time $t$, accounting for both cross-sectional and longitudinal coverage in a theoretically-grounded manner.
The post-hoc nature of \methodname facilitates its use as a general wrapper around any time series regression model.
We validate \methodname's performance through extensive experimentation: \methodname generally obtains efficient PIs and improves longitudinal coverage while preserving cross-sectional coverage.
\end{abstract}

\section{Introduction}
The impressive predictive performance of modern ``black-box'' machine learning methods has started to make them critical ingredients in various high-stakes decision-making pipelines.
It is thus increasingly important to quantify the predictive uncertainty of such models reliably and efficiently, which remains a fundamental challenge. 
Conformal Prediction (CP), pioneered by Vovk et al.~\cite{Vovk2005AlgorithmicWorld}, is a powerful framework for quantifying uncertainty under mild assumptions.
The model-agnostic and distribution-free nature of CP makes it particularly suitable for large neural network models, and has started to attract the attention of the deep learning community~\cite{angelopoulos2021uncertainty,Angelopoulos2022ImagetoImageRW, Bates2021DistributionFreeRP, CortsCiriano2019ConceptsAA,LVD, Zhang2021DeepLC}. 
The primary assumption in most current CP methods is that of data exchangeability. 
For instance, only assuming exchangeability of the calibration and test data, one can construct $1 - \alpha$ valid prediction intervals by simply querying the corresponding quantile of nonconformity scores on the calibration set.
Recent works have started exploring the adaptation of CP in settings that go beyond the usual exchangeability  assumption~\cite{barbercandes, ACI, Hu2020ADT,Oliveira2022SplitCP,Park2021PACPS, Podkopaev2021DistributionfreeUQ, Qiu2022DistributionfreePS, Tibshirani2019ConformalPU, Yang2022DoublyRC}, and to more complex data such as time series~\cite{ACI, alaaCFRNN, EnbPI_ICML, ACI_learn}.

We study the adaptation of CP to the cross-sectional time series regression setting. 
More formally, suppose our data comprises of $N$ time series, denoted $\{\mathbf{S}_i\}_{i=1}^{N}$, with each $\TS_i$ sampled from an arbitrary distribution $\mathcal{P}_S$. 
Further, each time series $\TS_i$ is a sequence of temporally-dependent random variables $[Z_{i,1} \ldots, Z_{i,t}, \ldots,Z_{i,T}]$
, with $Z_{i,t} = (X_{i,t} Y_{i,t})$ consisting of covariates $X_{\cdot, \cdot}\in \mathbb{R}^d$ and the response $Y_{\cdot,\cdot}\in\mathbb{R}$. 
Given data $\{Z_{N+1, j}\}_{j=1}^t$ until time $t$ for a new time series $\TS_{N+1}$, the time series regression problem entails predicting the response $Y_{N+1,t+1}$ at (unknown) time $t+1$. 
We are interested in quantifying the uncertainty of each prediction by constructing \emph{valid} prediction intervals (PI). 
That is, given a confidence level $1-\alpha$, we are interested in constructing a prediction interval, $\hat{C}_{\alpha, N+1,t+1}$, that will cover $Y_{N+1, t+1}$ with probability of at least $ 1-\alpha $. 

A crucial requirement in cross-sectional time series regression  is to distinguish two notions of validity, \emph{longitudinal} and \emph{cross-sectional}, to ensure reliable performance. 
Longitudinal validity is concerned with validity along the temporal axis for each time series. 
On the other hand, cross-sectional validity is concerned with validity \emph{across the populational cross-section} of the time series data. 
Figure~\ref{fig:main:validity} illustrates both notions of validity and a standard real-world occurrence of such a problem setting. 
Recently, several research groups have explored adapting CP to the time series setting. 
Some of this work~\cite{ACI,ACI_learn,EnbPI_ICML} focuses only on longitudinal validity, which is extremely difficult without strong distributional assumptions~\cite{barbercandes}. 
Furthermore, such methods cannot leverage rich information inherent in the cross-section. 
The only work which addresses cross-sectional validity is to due to Stankevi\v{c}i\=ut\.{e} et al.~\cite{alaaCFRNN}, but it ignores the temporal dependence. Accounting for both notions of coverage is critical to obtain reliable performance.

To remedy the above situation, we propose Temporal Quantile Adjustment (\methodname) for CP in the cross-sectional time series regression setting. 
\methodname is the \textit{first method} that can account for both cross-sectional and longitudinal validity simultaneously. 
Although \methodname can be used as a wrapper around any time series regression model, we focus on neural networks, as our main inspiration comes from complicated time series regression problems in healthcare\footnote{We include results for other models in the Appendix.}. 
Neural networks are particularly suited for such tasks, which can involve modeling the evolution of heterogeneous entities such as diagnostic and drug codes, patient and physician embeddings and regressing over a target of interest. 
Taking inspiration from~\cite{ACI}, \methodname adjusts the quantile to query at each time step in a theoretically-grounded manner. 
Based on the nature of quantile adjustment, we also propose two variants of \methodname, which further shed light on the generality of our method. 
The ability of \methodname to handle both cross-sectional and longitudinal validity is borne out in extensive experimentation, where it significantly outperforms competing methods.

\def \FigValidities{
\begin{figure}[t]
    \centering
    \includegraphics[width=0.95\textwidth]{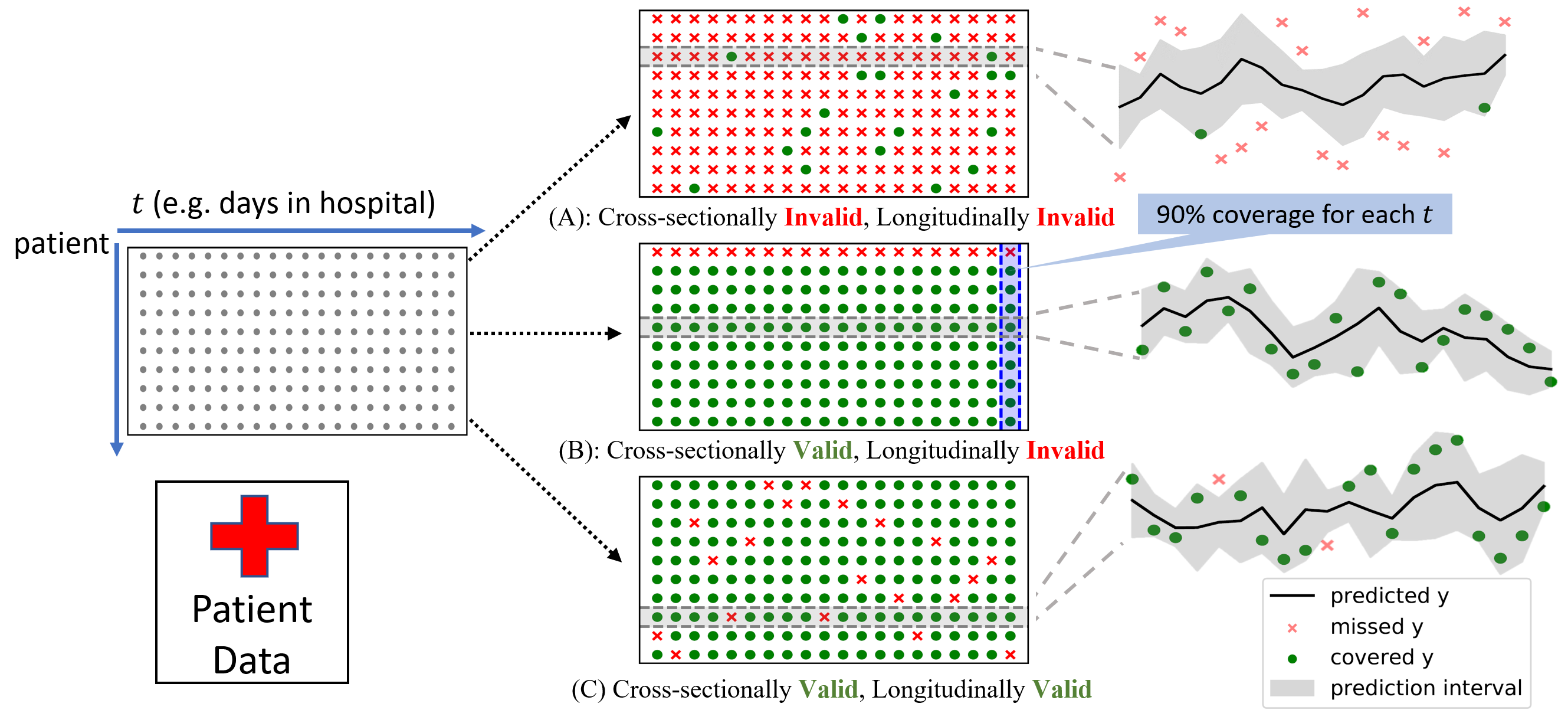}
\caption
{
PI estimators with different cross-sectional or longitudinal properties/validities.
\fontred{\texttimes} denotes the ground-truth $y$ that is outside of the PI. 
(A) features PIs that are not valid in either sense: $Y$ is never covered. 
(B) is cross-sectionally valid: for any $t$, 90\% of the $Y_{\cdot, t}$ are covered.
It is longitudinally \textit{invalid}, as 10\% of the TS receive no coverage at all. 
(C) exhibits both cross-sectional and longitudinal validity, which is ideal.
\label{fig:main:validity}
}
\vspace{-7.5mm}
\end{figure}
}
\FigValidities

\section{Related Work}

Our work falls squarely within the Conformal Prediction (CP) framework. 
The original formulation of CP was in a purely transductive setting~\cite{Saunders1999TransductionWC,Saunders2000ComputationallyET,Vovk2005AlgorithmicWorld}, and was computationally inefficient.
More efficient variants dubbed as Inductive Conformal Prediction (ICP)~\cite{Papadopoulos2008InductiveCP,Papadopoulos2002InductiveCM,Vovk2013ConditionalVO} were proposed soon after and were more broadly popularized by followup works in statistics~\cite{Lei2018Distribution-FreeRegression, Lei2013DistributionFreePS, lei2014}.
A similar idea, often referred to as Split Conformal~\cite{lei_conformal_2015}, is now more or less used interchangeably with ICP, and is fundamental to our paper.
The model-agnostic and distribution-free nature of split conformal makes it suitable for large black-box models, and thus it has seen adoption in deep learning-based pipelines (e.g.~\cite{angelopoulos2021uncertainty,Fannjiang2022ConformalPF,SplitDNN2, LVD, SplitDNN3,alaaCFRNN}). 

One of the mainstays of the CP framework is the assumption of exchangeability between calibration and test data. Extensions of CP ``beyond exchangeability'' have attracted relatively little attention until recently. A key publication in the area is~\cite{Tibshirani2019ConformalPU} which used the notion of weighted exchangeability to handle covariate shift. 
Several notable works that address various aspects of the covariate shift problem include~\cite{Hu2020ADT,Park2021PACPS, Yang2022DoublyRC, Qiu2022DistributionfreePS}. 
More recent work~\cite{barbercandes, ACI} handles gradual distribution drifts. \cite{barbercandes} additionally proposes an extension of CP when the data-points cannot be treated in a symmetric manner. Extensions of the split conformal method to a broad class of dependent processes such as stationary $\beta$-mixing processes was proposed by~\cite{Oliveira2022SplitCP}. These works provide some general methodological insights to model temporal dependence in our case, but are otherwise not directly related. In particular, the online adaptive method of~\cite{ACI}, served as a major source of inspiration for the development of a variant of \methodname (dubbed \methodError).
The most related works have a focus on generating valid intervals in time series regression~\cite{ACI,ACI_learn,barbercandes,EnbPI_ICML}. 
These works end up ignoring cross-sectional aspects, which is understandable given the tasks they study. 
On the other extreme is~\cite{alaaCFRNN}, which focuses only on cross-sectional coverage, ignoring the temporal dimension in constructing PIs.

Various other techniques, outside the ambit of CP, have also been extended to quantify uncertainty in time series forecasting as well. 
For instance, approximate Bayesian methods~\cite{HMC_Chen,Welling2011BayesianDynamics,NIPS1992_f29c21d4,VariationalBNN_Louizos,AutoEncodingVariationalBayes,Gal2016DropoutLearning,Lakshminarayanan2017SimpleEnsembles} are quite popular for uncertainty quantification, and have been extended to RNNs~\cite{BayesianRNN,BayesianRNNMisc1}.  
Finally, one may also use the idea of directly predicting the quantiles (as opposed to the point estimate) in regression tasks~\cite{Pinball1,PinballOldEcon}, and applying it to time series forecasting~\cite{QRNN,SplineQRNN}.
However, such methods usually require changing the base model and typically do not come with coverage guarantees.





\section{Preliminaries}\label{sec:prelim}
This section builds foundation for our exposition of \methodname. 
We begin by expounding further on longitudinal and cross-sectional validity, followed by presenting the   exchangeability assumption. Finally, we discuss the use of split conformal prediction to construct (cross-sectionally) valid PIs.



\subsection{Cross-sectional Validity vs Longitudinal Validity}\label{sec:prelim:xs_validity}

\textbf{Cross-sectional validity} is the more common type of validity encountered in CP, being the only type of validity in non-time-series settings. More formally:
\begin{definition}\label{def:valid:xs}
Prediction interval $\hat{C}_{\cdot,\cdot}$ is $1-\alpha$ cross-sectionally valid if, for any $t$,
\begin{align}
    \mathbb{P}_{\TS_{N+1}} \{Y_{N+1, t} \in \hat{C}_{N+1, t} \}\geq 1-\alpha.
\end{align}
\end{definition}
$\mathbb{P}_{\TS_{N+1}}$ means the probability is taken over the randomness of $\TS_{N+1}$.
If $\hat{C}_{N+1, t}$ is random (e.g. depends on $\{\TS_{i}\}_{i=1}^{N}$) then the probability is taken over the randomness of $\hat{C}_{N+1, t}$ as well.
Note that if we consider the case where every time series only consists of one step ($T=1$), then we recover the usual definition of marginal validity. 
Cross-sectional validity translates to high-probability in coverage for a \textit{randomly drawn time series}. As we will see later, cross-sectional validity is easier to achieve, since we can assume \textit{inter}-time-series exchangeability.


\textbf{Longitudinal validity}, on the other hand, is concerned with coverage along the temporal axis for a particular TS.
We use the following definition:
\begin{definition}\label{def:valid:long}
Prediction interval $\hat{C}_{\cdot, \cdot}$ is $1-\alpha$ longitudinally valid if for almost every time-series $\TS_{N+1}\sim \mathcal{P}_S$ there exists a $T_0$ such that:
\begin{align}
    t > T_0 \implies \mathbb{P}_{Y_{N+1, t}|\TS_{N+1,:t-1}} \{Y_{N+1, t} \in \hat{C}_{N+1,  t} \}\geq 1-\alpha.
\end{align}
\end{definition}
Here, the qualifier ``almost every'' means that the set of time series' for which such coverage may fail is of measure zero (under $\mathcal{P}_S$). 
The threshold $T_0$ allows some ``time'' for $\hat{C}$ to potentially adapt to the temporal information in a particular TS. 
It should be clear that longitudinal validity is harder to attain, because we can no longer marginalize the probability over randomly drawn time series. 

\subsection{Conformal Prediction for Cross-sectionally Valid PI}\label{sec:prelim:scp}
In this section we explain how to use conformal prediction to construct cross-sectionally valid PIs. 
We first introduce exchangeability assumption, which is the central assumption in conformal prediction, and slightly weaker than the standard i.i.d assumption. 
%
More formally:
\begin{definition}\label{def:exchangeable}(\textbf{Exchangeability}~\cite{Vovk2005AlgorithmicWorld}) 
A sequence of random variables, $Z_1, Z_2, \ldots, Z_n \in \mathcal{Z}$ are exchangeable if for any permutation $\pi: \{1,2,\dots,n\} \to \{1,2,\dots,n\}$, and every measurable set $E\subseteq \mathcal{Z}^n$, we have
\begin{align}
    \mathbb{P}\{(Z_1, Z_2,\ldots,Z_n) \in E\} = \mathbb{P}\{(Z_{\pi(1)}, Z_{\pi(2)},\ldots,Z_{\pi(n)}) \in E\}
\end{align}
\end{definition}

Definition~\ref{def:exchangeable} can naturally be extended to a sequence of randomly drawn time series:
\begin{definition} \label{def:exchangeable:ts} (\textbf{Exchangeable Time Series}) 
Given time series $\TS_1, \TS_2, \ldots, \TS_n$ where $\TS_i = [Z_{i,1},\ldots,Z_{i,T},\ldots]$, denote $Z_{i, \{t_j\}_{j=1}^m}$ as the concatenated random variable of $(Z_{i, t_1}, \ldots, Z_{i, t_m})$.
Time series $\TS_1, \TS_2, \ldots, \TS_n$ are exchangeable if, for any finitely many $t_1 < \cdots < t_m$, the random variables $Z_{1, \{t_j\}_{j=1}^m},\ldots,Z_{n, \{t_j\}_{j=1}^m}$ are exchangeable.
\end{definition}
As a concrete example, suppose we \textit{randomly} pick 100 patients from a hospital's EHR database for predicting readmission risk. It is fairly reasonable to assume that these time series are exchangeable, despite the obvious strong temporal dependence \textit{within} each time series.
\textit{Throughout this paper, we will assume $\TS_1,\ldots,\TS_{N+1}$ are exchangeable time series.}

We now explain how to construct cross-sectionally valid PIs.
To construct a PI for $Y_{i,t}$, we first split our training data $\{\TS_i\}_{i=1}^N$ into a \textit{proper training set} and a \textit{calibration set}~\cite{Papadopoulos02}.
The training set is used to train models for the \textit{nonconformity score} function, and the calibration set is used to collect such nonconformity scores (denoted as $V(\cdot)$). 
For example, one may train a mean estimator $\hat{\mu}$ (e.g. an RNN) on the training set, and use the absolute residual $v_{i,t} \gets |y_{i,t} - \hat{y}_{i,t}|$,  where $\hat{y}_{i,t} = \hat{\mu}(X_{i,t};\TS_{i, :t-1})$, as the nonconformity score. 
Here $\TS_{i, :t-1}$ denotes $[Z_{i,1},\ldots,Z_{i,t-1}]$.
The idea behind the split conformal method is that the scoring function (e.g. $\hat{\mu}$) is only fit on the proper training set, implying that nonconformity scores on the calibration set and $v_{N+1,t}$ are also \textit{exchangeable}. Here on, for the sake of simplicity of notation, we will use $\{\TS_{i}\}_{i=1}^{N}$ to denote the \textit{calibration set only}, assuming all necessary models have already been trained.

Given a nonconformity score $V(\cdot)$ (possibly using some trained model $\hat{\mu}$) and a set of exchangeable time series $\{\TS_i\}_{i=1}^{N+1}$, the split conformal method can be used to generate the following $1-\alpha$ cross-sectionally valid prediction interval:
{\small
\begin{align}
    \hat{C}^{\methodSplit}_{N+1,t+1} = \Bigg\{y: V_{N+1,t+1}(\hat{y}, y) \leq Q\Big(1-\alpha; \{v_{j,t+1}\}_{j=1}^{N}\cup \{V_{N+1,t+1}(\hat{y}, y)\}\Big)\Bigg\}.
\end{align}
}
Here, $Q(\beta; A)$ means the $\beta$-quantile for the set $A$.
As a concrete example, if we let $v_{i,t} = |y_{i,t} - \hat{y}_{i,t}|$, and employ the standard trick that replaces $v_{N+1,t+1}$ with $\infty$ to avoid plugging in (uncountably) many values for $y$~\cite{barbercandes}, 
the prediction interval for $Y_{N+1, t+1}$ becomes:
{\small
\begin{align}
    \hat{C}^{\methodSplit}_{N+1,t+1} &\defeq [\hat{y} - \hat{v}, \hat{y} + \hat{v}] 
    \text{ where } \hat{v} \defeq Q\Big(1-\alpha; \{|y_{i,t+1} - \hat{y}_{i,t+1}|\}_{i=1}^N \cup \{\infty\}\}\Big)
\end{align}
}
Assuming exchangeability, we can easily show the cross-sectional validity of $\hat{C}^{\methodSplit}$:
\begin{theorem}\label{thm:cp:valid}
(\cite{barbercandes,Vovk2005AlgorithmicWorld})
$\hat{C}^{\methodSplit}$ is $1-\alpha$ cross-sectionally valid (Def.~\ref{def:valid:xs}).
\end{theorem}
The intuition behind the proof is that the exchangeability of the time series translates to the exchangeability of the nonconformity scores, which means the rank of $v_{N+1, t+1}$ among $\{v_{i,t+1}\}_{i=1}^{N+1}$ follows a uniform distribution. The coverage guarantee in Theorem~\ref{thm:cp:valid} then follows. 
Note that with a finite calibration set, the $1-\alpha$-quantile could be ambiguously defined, and in practice one would use $\frac{\ceil{(1-\alpha)(N+1)}}{N+1}$ to get a slightly more conservative PI, or ``flip a (biased) coin'' to choose between $\frac{\ceil{(1-\alpha)(N+1)}}{N+1}$ and $\frac{\floor{(1-\alpha)(N+1)}}{N+1}$ for a precise $1-\alpha$ coverage (e.g. the ``smoothed'' ICP in~\cite{Vovk2005AlgorithmicWorld} 
or the tie-breaking trick in~\cite{angelopoulos2021uncertainty}). 
In Section~\ref{sec:method}, we  assume the precise coverage PI for the ease of discussion.

While cross-sectionally valid, $\hat{C}^{\methodSplit}$ ignores the temporal dependence in the nonconformity scores completely.
We will explain in Section~\ref{sec:method} how to adapt to the temporal dependence by ``quantile adjustment'', thus improving longitudinal coverage as well.


%
\section{Temporal Quantile Adjustment (\methodname)}\label{sec:method}
In Section~\ref{sec:prelim} we discussed a classical conformal prediction method and also highlighted its inherent limitations in the time series setting.  
In this section we will formally introduce Temporal Quantile Adjustment (\methodname), which queries quantiles differently than in the aforementioned split conformal procedures.
We first explicate the goals and motivations of \methodname in Section~\ref{sec:method:goals}.
Then, in Section~\ref{sec:method:budget} and \ref{sec:method:error} we propose two principled adjustment methods along with theoretical analyses.

\subsection{Improving Longitudinal Coverage}\label{sec:method:goals}

Although it is tempting to directly pursue distribution-free finite-sample PI estimator that achieves longitudinal \textit{coverage guarantee} (Def.~\ref{def:valid:long}), it is likely too optimistic due to the fact that $[Z_{i,1},\ldots,Z_{i,T}]$ are not exchangeable and we cannot characterize them meaningfully without imposing (strong) distributional assumptions. For example, in~\cite{barbercandes}, the bound for coverage gap\textemdash which captures the loss in coverage compared to what is achievable under exchangeability\textemdash essentially becomes 1 as the
data becomes time-dependent. 
In fact, one could view Def.~\ref{def:valid:long} as a ``conditional validity'' (where $\TS_{N+1,:t}$ as a whole is viewed as the input), and it is well-known that distribution-free finite-sample conditional validity is impossible to achieve in a meaningful way~\cite{barber2020limits,lei2014}.

Nevertheless, longitudinal coverage may still be empirically \textit{improved} if we can adapt to the temporal dependence. 
From our EHR example, suppose $\hat{\mu}$ has been giving high-error predictions for a patient for 8 out of the past 10 days, we might suspect high error going forward for this patient as well.
If this is indeed the case, then naively applying $\hat{C}^{\methodSplit}$ can only attain low coverage for this patient, no matter how long a history we observe. 
To address this, we propose to query different quantiles based on the partially observed time series. 

From now on, we use $\hat{C}_{a_{i,t}, i, t}$ to denote a PI for $Y_{i,t}$ with a pre-specified target coverage of $1-a_{i,t}$, in order to emphasize dependence on the ``quantile to query''.
We let $a_{i,t} = \alpha - \hat{\delta}_{i, t}$, where $\hat{\delta}_{i, t}$ is the \textit{quantile adjustment}.
Classical split conformal PIs (e.g.~\cite{alaaCFRNN}) entail a special case: $\forall i, \forall t, \hat{\delta}_{i,t} \equiv 0$.
We refer to this method as Temporal Quantile Adjustment (\methodname). 

Now, denote the random variable $R_{i, t}$ as the rank/quantile of $V_{i,t}$ among $\{V_{j,t}\}_{j=1}^{N+1}$:
\begin{align}
    r_{i,t} \defeq Q^{-1}(v_{i,t}; \{v_{j,t}\}_{j=1}^{N+1}) \defeq \frac{|\{j: v_{j,t} < v_{i,t}\}|}{N+1}.
\end{align}
For example, if $v_{i,t}$ is the smallest among $\{v_{j,t}\}_{j=1}^{N+1}$, then $r_{i,t} = 0$.
Intuitively, we would like to use a more conservative (smaller) $a_{N+1, t}$ when we believe $\TS_{N+1}$ as a whole is less ``conformal'' and $R_{N+1,t}$ is likely high. 
A crucial observation is that if there is no actual temporal dependence between the nonconformity scores (i.e. we are just adjusting $a_t$ based on some ``noise''), then we \textit{do not lose any coverage} as long as the expected adjustment is zero.
\begin{theorem}\label{thm:a_t:noworse}
If the nonconformity score's rank $(R_{N+1,t})$ is independent of the quantile adjustment $(\hat{\delta}_{N+1,t})$, then
$\mathbb{P}_{\TS_{N+1}}\{Y_{N+1, t} \in \hat{C}_{a_{N+1,t}, N+1, t}\} \geq 1-\alpha + \mathbb{E} [\hat{\delta}_{N+1,t}]$.
\end{theorem}
All proofs are deferred to the Appendix.

\textbf{Remark}:
The assumption in Theorem~\ref{thm:a_t:noworse} is \textit{not} that $\TS_{N+1}$ itself is not temporally dependent, nor is it the slightly weaker assumption, that of the temporal independence of either $\{R_{\cdot, t}\}_{t=1}^T$ or $\{V_{\cdot, t}\}_{t=1}^T$. The assumption in Theorem~\ref{thm:a_t:noworse} only suggests that there is no temporal pattern in the prediction errors that $\hat{\delta}$ can capture.
This could happen, for example, when the RNN captures the underlying data generating process fully, but only misses the random noise (aleatoric uncertainty).
This could also happen if our quantile adjustments $(\hat{\delta})$ are pure noise.

Unfortunately, although it can be tempting to conclude that $\mathbb{E} [\hat{\delta}_{N+1,t}]=0$ implies finite-sample cross-sectional validity, this conclusion would be incorrect, because it ignores the dependence between $\mathbf{1}\{Y_{N+1,t}\in\hat{C}_{a_{N+1,t}, N+1,t}\}$ and $a_{N+1, t}$. 
However, we will next discuss how to perform quantile adjustment, and why it typically \textit{improves} coverage.

\subsection{Quantile Budgeting (\methodBudget)}\label{sec:method:budget}

\def \FigTQA{
\begin{figure}[t]
    \centering
    \includegraphics[width=1\textwidth]{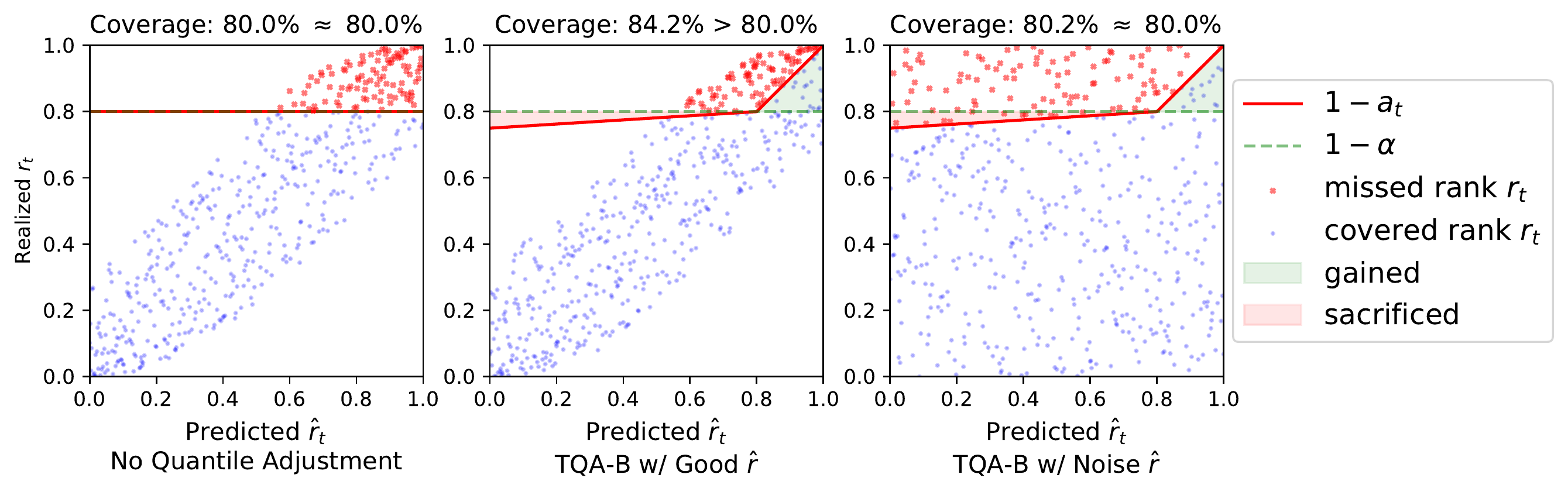}
\caption{
Coverage profiles with hypothetical realized rank $r$ condition on prediction $\hat{r}$, with $\alpha=0.2$ for readability.
($Y_{i,t} \in \hat{C}_{i,t} \Leftrightarrow r_{i,t} \leq 1-a_{i,t}$.)
As $\hat{r}$ follows a uniform distribution, the proportion of dots below the {\color{red} red line} 
represents the cross-sectional coverage probability.
\methodBudget generally improves coverage if $\hat{r}$ is correlated with the realized $r$ (middle), and does not lose coverage otherwise (right).
``Budgeting'' refers the constraint that  \colorbox{red!15}{sacrificed} and \colorbox{green!15}{gained} have equal areas.
\label{main:fig:tqa}
}
\vspace{-4mm} 
\end{figure}
}
\FigTQA

Theorem~\ref{thm:a_t:noworse} provides an interesting constraint that we should consider while designing $\hat{\delta}$. That is, we should let $\mathbb{E}[\hat{\delta}] = 0$ so as to keep the same coverage when we \textit{cannot} predict the quantiles, but hopefully improving coverage when we can. This suggests a design of $\hat{\delta}$ that we refer to as a type of ``budgeting''. Although it is possible to directly predict a good $\hat{\delta}_{i,t+1}$, we adopt a more principled two-step approach:
\begin{enumerate}[label=(\roman*)]
    \item We predict quantile $\hat{r}_{i,t+1}$ which estimates $r_{i, t+1}$.
    \item We use a pre-defined mapping $g$ to define the quantile adjustment $\hat{\delta}_{i,t+1}\gets g(\hat{r}_{i,t+1}; \alpha)$.
\end{enumerate}
  
This also permits research to improve each component independently. 
We introduce one alternative for each step in the Appendix.

\textbf{(i) Quantile Prediction}: 
The quantile prediction $\hat{r}_{i, t+1}$ is estimated by a function of the form $f(\TS_{i,:t};\{\TS_{j,:t}\}_{j=1}^{N+1})$.
Since $\hat{r}_{i,t+1}$ is supposed to predict $r_{i,t+1}$, \textit{the rank} of the nonconformity score, we impose the constraint that $\hat{r}_{i,t+1}$ should follow a uniform distribution over $\{\frac{j}{N}\}_{j=0}^{N}$.
We will focus on a simple rank prediction method stated below:
{\small
\begin{align}
    \hat{r}^{ms}_{i,t+1} \defeq Q^{-1}(\overline{\epsilon}_{i,t}; \{\overline{\epsilon}_{j,t}\}_{j=1}^{N+1}) & \text{ where }  \overline{\epsilon}_{i,t}  \defeq\sum_{t'=1}^{t} \frac{|y_{i,t'} - \hat{y}_{i,t'}|}{t}\beta^{(t-t')}.
\end{align}
}
Here, $\overline{\epsilon}_{i,t}$ is the exponentially decayed mean residual of time series $i$ up to time $t$, and we use $\beta=0.8$. 
Note that taking the rank with $Q^{-1}$ achieves the uniformity requirement, and $\overline{\epsilon}_{\cdot,t}$ could be replaced by any scoring function that takes into account the temporal information.

\textbf{(ii) Budgeting}:
Given prediction $\hat{r}_{i,t}$, we propose the following adjustment $\hat{\delta}_{i,t} \defeq g^{\methodBudgetSuffix}(\hat{r}_{i,t};\alpha)$:
{\small
\begin{align}
    g^{\methodBudgetSuffix}(r;\alpha) \defeq
    \begin{cases}  C(r - (1-\alpha)) & (r < 1-\alpha)  \\
    (r - (1-\alpha)) & (r \geq 1-\alpha)  \\ 
    \end{cases}  
    \text{ where } 
    C = \frac{(2\alpha N -\floor{\alpha N})(\floor{\alpha N} + 1)}{\ceil{(1-\alpha) N}((1-2\alpha) N + 1 + \floor{\alpha N})}.
\end{align}
}

The particular coefficient design used ensures that the following property holds:
\begin{theorem}\label{thm:budget:expectation}
Using $\hat{r}^{ms}_{\cdot,\cdot}$ and $g^{\methodBudgetSuffix}$, 
$\forall t, \mathbb{E}_{\TS_{N+1}}[a^{\methodBudget}_{N+1,t}] =  \alpha$.
\end{theorem}
Thus, by Theorem~\ref{thm:a_t:noworse}, \methodBudget will not lose coverage if $\hat{r}$ and $r$ are independent.
The following theorem provides a worst-case cross-sectional coverage guarantee, regardless of how ``bad'' $\hat{r}$ is:
\begin{theorem}\label{thm:budget:guarantee}
$\mathbb{P}_{\TS_{N+1}\sim\mathcal{P}_S}\{ Y_{N+1,t+1}\in\hat{C}^{\methodBudget}_{\alpha, N+1,t+1} \} \geq 1-\alpha - \underbrace{\Big(\frac{\alpha + \frac{1}{2N}}{1-\alpha  + \frac{1}{2N}}\Big)^2 (1-\alpha)}_{\text{worst-case loss}}.$
\end{theorem}
The worst-case loss term is typically small: about $0.012$ for $\alpha=0.1$ and $N = 100$, although it can also be also be high for a large $\alpha$ like $0.5$.
In practice, it is unlikely that $\hat{r}$ is worse than a random guess; the coverage is typically greater than $1-\alpha$, as we will see in the experiments and illustrated in Figure~\ref{main:fig:tqa}. 
In the Appendix, we present a more aggressive quantile adjustment function $g$ that provides a weaker guarantee than Theorem~\ref{thm:budget:guarantee}, but empirically performs better.

\textbf{Implementation Details}
To avoid creating infinitely-wide PIs, we could also let $\hat{\delta} = \lambda g(\hat{r};\alpha)$ so $a_t$ is bounded away from 0 (In our experiments we choose $\lambda$ such that $a_t \geq 0.01$). 
Moreover, the specific form of $C$ presented in this section depends on the concrete distribution of $\hat{r}$.
For example, $C$ would take a different form if $\hat{r}$ is defined to be uniform over $\{\frac{j}{N+1}\}_{j=1}^{N+1}$ rather than $\{\frac{j}{N}\}_{j=0}^N$.
Practically, we can simply let $C = \alpha^2(1-\alpha)^{-2}$ regardless of $N$.
The additional loss in Theorem~\ref{thm:budget:guarantee} will become $\alpha^2/(1-\alpha)$, and the change in Theorem~\ref{thm:budget:expectation} is negligible for a reasonable value of $N$.

\subsection{Error-based adjustment (\methodError)}\label{sec:method:error}
Another simple quantile adjustment approach is using a heuristic that depends on the past ``errors'':
Define $err_t = \mathbf{1}\{Y_t \not\in\hat{C}_{a_t}\}$, and increase $\delta_{t+1}$ (conservative) if we see too many errors in $\{err_{t'}\}_{t'<t+1}$ compared with $\alpha$, and vice versa.
Since this approach does not depend on the cross-section, we drop the subscript $\cdot_{N+1}$ for simplicity.
We use the following update rule (with $\hat{\delta}_0 = 0$) inspired by~\cite{ACI}\footnote{
Despite the similarity on the surface, \cite{ACI}  has no notion of cross-section.
}:
{\small
\begin{align}
    \hat{\delta}_{t+1} \gets  \begin{cases}
    \hat{\delta}_t + \gamma (err_t - \alpha) & (\hat{\delta}_t \geq \alpha-1) \\
    (1- \gamma)\hat{\delta}_t & (otherwise)
    \end{cases}
    \label{eq:a_t:update}.
\end{align}
}
Note that we do not explicitly impose the restriction that $\alpha-\hat{\delta}_t =a_t\in[0,1]$, which means the PI could have \textit{infinite width}. 
However, infinite-wide PI means no error, so $\hat{\delta}_{t+1}$ will decrease and we resume to a finite PI gradually.

As $\hat{\delta}_{t}$ depends on the entire error history, it is not immediately clear whether \methodError is still valid with the assumption in Theorem~\ref{thm:a_t:noworse}.
Below we state a ``no-worse'' type theorem for \methodError:
\begin{theorem}\label{thm:aci:finite}
If we assume the nonconformity score's rank has no temporal dependence, then
\begin{align}
    \forall t, \mathbb{E}_{\TS_{N+1}\sim\mathcal{P}_S}[a^{\methodError}_{N+1,t}] \leq \alpha.
\end{align}
Thus, $\hat{C}^{\methodError}$ is finite-sample cross-sectional $1-\alpha$ valid following Theorem~\ref{thm:a_t:noworse}.
\end{theorem}

Finally, an asymptotic longitudinal validity result 
can also be shown for long time series\footnote{
The update rule as shown in Eq.~\ref{eq:a_t:update} creates an asymmetry to accommodate for the asymptotic guarantee in Theorem~\ref{thm:aci:asymp}, which is why the expectation in Theorem~\ref{thm:aci:finite} is not an equality but an inequality. 
If asymptotic coverage is not a concern (because $T$ is small), 
one could use $(1- \gamma)\hat{\delta}_t$ when $\alpha-\hat{\delta}_t =a_t< 0$ as well, 
in which case Theorem~\ref{thm:aci:finite} becomes an equality, as we discuss in the proof for Theorem~\ref{thm:aci:finite} in the Appendix.
In practice (as we observe in our experiments), the difference in behavior is negligible. 
}:
\begin{theorem}\label{thm:aci:asymp} (Asymptotic Longitudinal Coverage)
For any time series $\TS$, 
$    \lim_{T\to\infty} \frac{\sum_{t=0}^{T-1} err_t}{T} \leq \alpha$.
\end{theorem}

\textbf{Remarks}:
Although Theorem~\ref{thm:aci:asymp} seems to suggest some sort of longitudinal validity, it does not contradict the hardness claim in Section~\ref{sec:method:goals}, because \methodError achieves this via infinitely-wide PIs.
We also refer interested readers to~\cite{ACI_learn} as an example of an different heuristic based on errors of a single time series. 
However, it will require further modifications for finite-sample cross-sectional validity.

\section{Experiments}\label{sec:exp}
In this section, our goal is to to verify the following empirically:
\begin{enumerate}
    \item \methodname maintains cross-sectional coverage and achieves competitive PI efficiency.
    \item Ignoring temporal dependence in naive split conformal prediction leads to low longitudinal coverage for some TS.
    \item \methodname improves longitudinal coverage.
    
\end{enumerate}

\textbf{Baselines}:
We use the following state-of-the-art baselines for PI construction:
Conformal forecasting RNN (\baselineCFRNN)~\cite{alaaCFRNN}), a direct application of split-conformal prediction~\cite{Vovk2005AlgorithmicWorld}\footnote{
\cite{alaaCFRNN} suggests performing Bonferroni correction to jointly cover the entire horizon (all $T$ steps).
We explain in the Appendix why this is problematic. 
};
Quantile RNN (\baselineQRNN)~\cite{QRNN}, which directly predicts the two endpoints (represented by two quantiles) of the PI;
RNN with Monte-Carlo Dropout (DP-RNN)~\cite{Gal2016DropoutLearning};
Conformalized Quantile Regression with \baselineQRNN (\baselineCQR)~\cite{CQR_NEURIPS2019_5103c358}, which, as the name suggests is a conformalized version of quantile regression;
Locally adaptive split conformal prediction (\baselineMADSplit)~\cite{Lei2018Distribution-FreeRegression}, which uses a normalized absolute error as the nonconformity score (we follow the implementation in~\cite{CQR_NEURIPS2019_5103c358}).

\def \TabDataSumm{
\begin{table}[ht]
\vspace{-4mm}
\captionof{table}{
Number of TSs in each dataset along with the length.
}
\centering
\begin{small}
\setlength\tabcolsep{6pt}
\begin{tabular}{l|lllll}
\toprule
Properties & \dataMIMIC & \dataCLAIM & \dataCOVID & \dataEEG & \dataGEFCom/\dataGEFComR \\
\midrule
\# train/cal/test  & 192/100/100 & 2393/500/500 &  200/100/80 & 300/100/200 & 1198/200/700 \\
$T$ (length)   & 30 & 30 & 30 & 63 & 24 \\
\bottomrule
\end{tabular}
\label{table:exp:summ}
\end{small}
\vspace{-2mm}
\end{table}
}
\TabDataSumm

\def \WrapFigCov{
\begin{wrapfigure}[21]{L}{0.44\textwidth}
\begin{minipage}{0.42\textwidth}
\vspace{-2mm}
\includegraphics[width=1\textwidth]{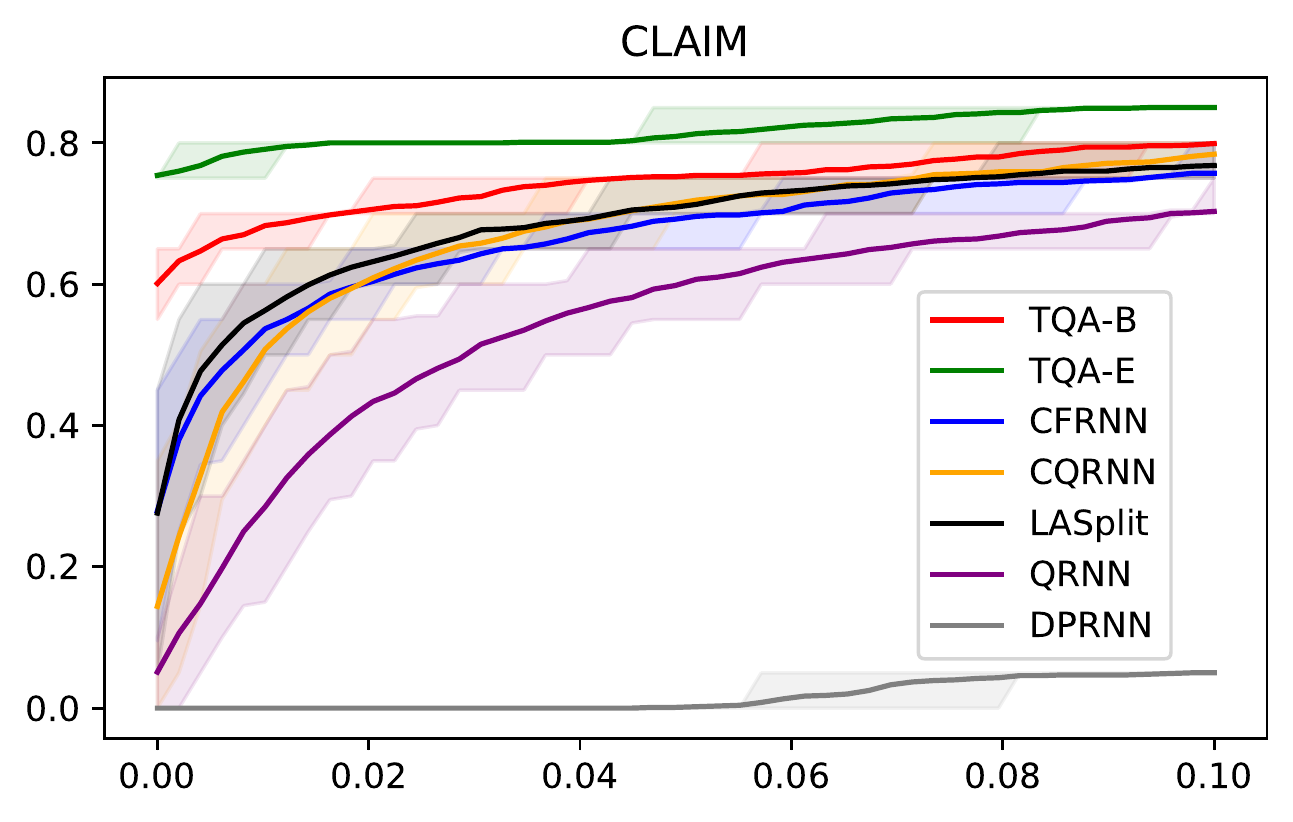}
\caption{
  Coverage rate (Y-axis) vs the percentile among all test TS (X-axis, with zero meaning the least-covered TS) for the 10\% least-covered TS in \dataCLAIM.
  The bands denote the center 80\% realizations.
  \methodError has an natural advantage 
  by using infinitely-wide PIs.
  However, even \methodBudget still significantly improves the longitudinal coverage rate over all baselines.
}
\label{wrapfig:exp:tail}
\vspace{-2mm}
\end{minipage}
\end{wrapfigure}
}

\def \WrapFigCovResid{
\begin{wrapfigure}[36]{L}{0.43\textwidth}
\begin{minipage}{0.42\textwidth}
\vspace{-7mm}
    \includegraphics[width=1\textwidth]{NeurIPS_TCR_CLAIM.pdf}
  \caption{
  Coverage rate (Y-axis) vs the percentile among all test TS (X-axis, with zero meaning the least-covered TS) for the 10\% least-covered TS in \dataCLAIM.
  The bands denote the center 80\% realizations.
  \methodError has an natural advantage 
  by using infinitely-wide PIs.
  However, even \methodBudget still significantly improves the longitudinal coverage rate over all baselines.
}
\label{wrapfig:exp:tail}
  \includegraphics[width=1\textwidth]{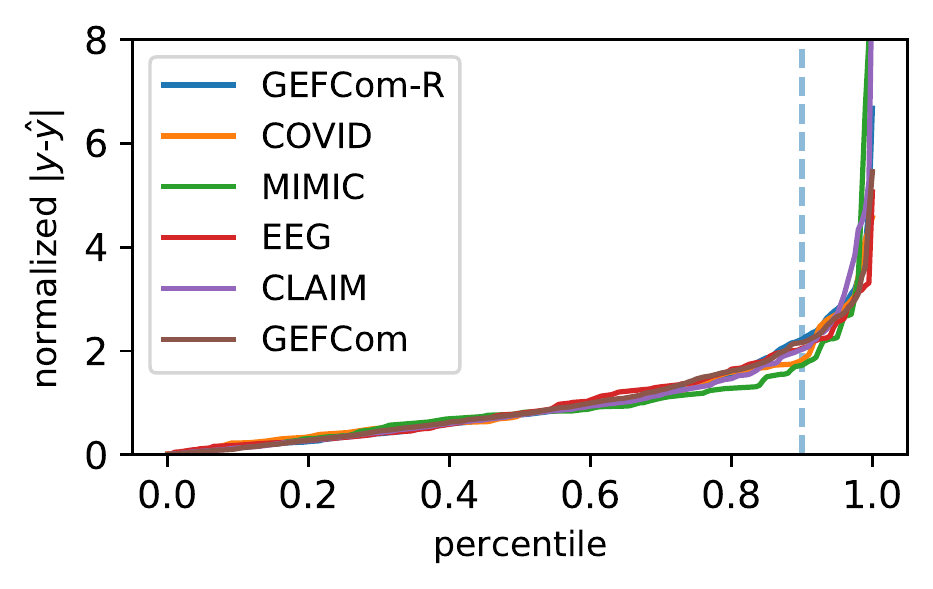}
  \caption{
  Sorted absolute residuals ($|y-\hat{y}|$) 
  for $t=T-10$. 
Each dataset is normalized so the mean of the residuals is 1.
To cover extreme values, even if allowed to ``sacrifice'' some less extreme values, the PI on average is expected to get much wider.
It is thus surprising that \methodBudget could improve \textit{both} efficiency and the tail coverage.
}
\label{wrapfig:exp:resid}
\vspace{-2mm}
\end{minipage}
\end{wrapfigure}
}
\WrapFigCovResid

\textbf{Datasets}
We test our methods and baselines on the following datasets:
Electronic health records data for white blood cell counts (WBCC) prediction (\dataMIMIC ~\cite{MIMIC,PhysioNet,MIMICDemo}),
COVID-19 cases prediction (\dataCOVID~\cite{COVID}),
Electroencephalography trajectory prediction after visual stimuli (\dataEEG~\cite{UCI_EEG}),
energy load forecasting (\dataGEFCom~\cite{GEFCom}),
and healthcare claim amount prediction (\dataCLAIM) using data from a large American healthcare data provider.
Among these, we mostly follow~\cite{alaaCFRNN} in preparing \dataMIMIC, \dataCOVID and \dataEEG.
Note that \dataGEFCom is originally a single time series (hourly observations for years).
Therefore, we treat each day as a single TS, and perform a strict temporal splitting (test data is preceded by calibration data, which is preceded by the training data), which means \textit{exchangeability is broken}.
We also include a \dataGEFComR(andom) version that preserves the exchangeability by ignoring the temporal order in data splitting.
Table~\ref{table:exp:summ} provides a brief summary of the data.
Due to space constraints, the details for each dataset are relegated to the Appendix.

\def \WrapFigResid{
\begin{wrapfigure}[18]{L}{0.44\textwidth}
\begin{minipage}{0.42\textwidth}
\vspace{-7mm}
\includegraphics[width=1\textwidth]{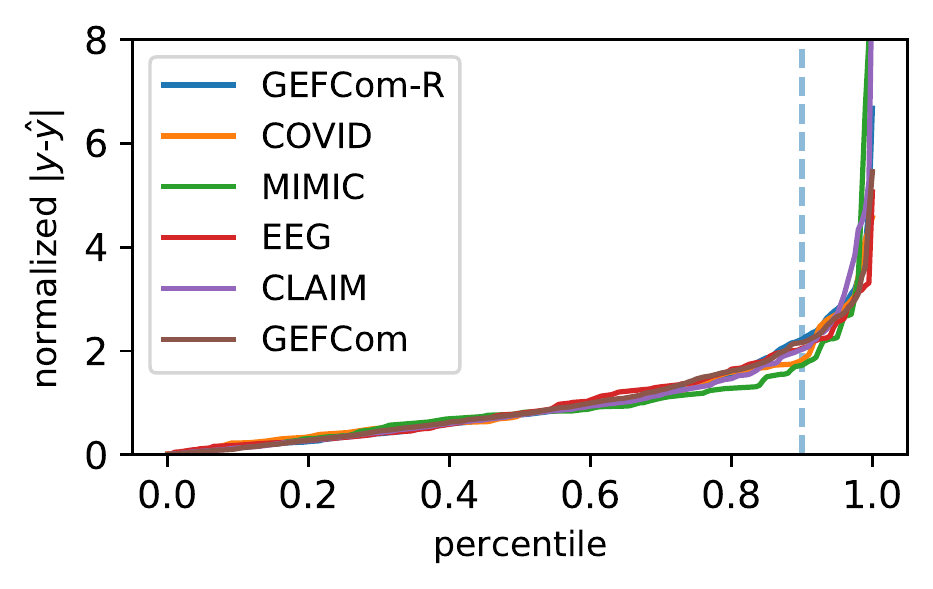}
  \caption{
  Sorted absolute residuals ($|y-\hat{y}|$) 
  for $t=T-10$. 
Each dataset is normalized so the mean of the residuals is 1.
To cover extreme values, even if allowed to ``sacrifice'' some less extreme values, the PI on average is expected to get much wider.
It is thus surprising that \methodBudget could improve \textit{both} efficiency and the tail coverage.
}
\label{wrapfig:exp:resid}
\end{minipage}
\end{wrapfigure}
}

\textbf{Evaluation Metrics and Experiment Setup}
We use RNN as the base point estimator due to its flexibility and for comparison with~\cite{alaaCFRNN}. 
We use $\alpha=0.1$, and a LSTM~(\cite{LSTM}) similar to that in~\cite{alaaCFRNN} (full implementation details in the Appendix).
For \methodError, we use $\gamma=0.005$ following~\cite{ACI}. 
For each dataset, we repeat the prediction task 50 times, and report the mean and standard deviation of the average coverage rate, tail coverage rate, and inverse coverage efficiency for the last 20 time-steps. 
Here, tail coverage rate means the average coverage rate of the least-covered 10\% of the time series.
A high tail coverage rate thus implies better longitudinal coverage.
Inverse coverage efficiency is measured by the average PI width divided by the marginal coverage rate (the smaller the better).
Since \methodError could create infinite PIs, we replace $\infty$ with 2x the widest finite PI.
We also include in the Appendix results on full time series, and with Linear Regression instead of LSTM to show the model-agnostic nature of \methodname.

Note that although we use equal-length time series in these experiments, for data with variable lengths (such as \dataCLAIM or \dataMIMIC), one could filter the calibration set before querying the quantile. 
As long as we assume exchangeability conditioning on     length, all theoretical analysis still holds.

\textbf{Results}
The results are presented in Tables~\ref{table:exp:real:coverage}, \ref{table:exp:real:tcr}, and~\ref{table:exp:real:inveff}.
In Table~\ref{table:exp:real:coverage}, we verify that conformal prediction methods - \baselineCFRNN, \baselineCQR, \baselineMADSplit and \methodname - are empirically cross-sectionally valid. 
The non-conformal methods (\baselineQRNN and \baselineDPRNN) have unreliable coverage.
In Table~\ref{table:exp:real:tcr}, we show that \methodname can greatly improve the (longitudinal) average coverage rate for the worst TS.
(\methodError is consistently better than \methodBudget due to the presence of infinitely-wide PIs.)
This is also visualized in Figure~\ref{wrapfig:exp:tail}.
Note that although \baselineCQR and \baselineMADSplit do not perform quantile adjustment, they model uncertainty directly, which also helps improve the longitudinal coverage but is less robust.
In Table~\ref{table:exp:real:inveff}, we verify that \methodname did not achieve better coverage simply by using very wide PIs (which is however the case for \baselineDPRNN on \dataGEFCom). 
This is somewhat surprising because from Figure~\ref{wrapfig:exp:resid}, the marginal gain in coverage decreases fast as $a_t$ decreases.
The PIs for \methodBudget should be wider due to the slight over-coverage, and the convexity (with any quantile adjustment).
This suggests that \methodBudget performs the budgeting very efficiently to cancel out both effects.
The efficiency of \methodError seems low due to the infinitely-wide PIs (replaced by 2x maximum finite width in this computation), but we will see that it generates mostly finite PIs, and the median width is still competitive.

Finally, we would like to also emphasize that any nonconformity scores could theoretically be combined with \methodname.
In this paper, and in our experiments, we mostly tried to combine the simplest nonconformity scores used in \baselineCFRNN with \methodname.
The question of how to combine \methodname with other nonconformity scores (such as those used in \baselineCQR or \baselineMADSplit) is left for future research.


\begin{table}[ht]
\vspace{-3mm}
\captionof{table}{
Average coverage rate. 
Empirically valid methods are in \textbf{bold} (at p = 0.01).
As expected, conformal baselines are valid, while others (\baselineQRNN and \baselineDPRNN) are not.
Note that \dataGEFCom does not satisfy the exchangeability assumption, causing invalid coverage for most conformal methods.
However, \methodname still outperforms all conformal baselines, with \methodError still valid. 
}
\centering
\begin{scriptsize}
\setlength\tabcolsep{6pt}
\begin{tabular}{l|ll|lll|ll}
\toprule
Coverage       &\methodBudget & \methodError & \baselineCFRNN & \baselineCQR & \baselineMADSplit &  \baselineQRNN  & \baselineDPRNN  \\
\midrule
\dataMIMIC    &  \textbf{91.31$\pm$1.32} & \textbf{91.19$\pm$0.48} & \textbf{90.06$\pm$1.73} & \textbf{90.15$\pm$1.24} & \textbf{90.33$\pm$1.54} & 86.90$\pm$1.22 & 46.30$\pm$3.84\\
\dataCLAIM & \textbf{91.10$\pm$0.49} & \textbf{91.56$\pm$0.35} & \textbf{90.21$\pm$0.56} & \textbf{90.15$\pm$0.68} & \textbf{90.20$\pm$0.64} & 85.90$\pm$0.78 & 24.79$\pm$0.85\\
\dataCOVID & \textbf{90.79$\pm$1.45} & \textbf{91.73$\pm$0.85} & \textbf{90.25$\pm$1.69} & \textbf{90.08$\pm$1.62} & \textbf{90.18$\pm$1.46} & 89.19$\pm$1.54 & 67.51$\pm$3.76\\
\dataEEG & \textbf{90.73$\pm$1.21} & \textbf{90.63$\pm$0.75} & \textbf{89.92$\pm$1.44} & \textbf{89.99$\pm$1.76} & \textbf{89.80$\pm$1.15} & 87.96$\pm$0.82 & 39.24$\pm$1.30\\
\dataGEFCom & 89.58$\pm$0.25 & \textbf{90.94$\pm$0.14} & 88.61$\pm$0.16 & 89.16$\pm$0.17 & 88.96$\pm$0.18 & 80.40$\pm$1.36 & 89.50$\pm$0.73\\
\dataGEFComR     & \textbf{90.56$\pm$0.64} & \textbf{90.72$\pm$0.45} & \textbf{89.92$\pm$0.78} & \textbf{90.07$\pm$0.63} & \textbf{89.95$\pm$0.72} & 85.49$\pm$1.08 & \textbf{91.03$\pm$0.76} \\
\bottomrule
\end{tabular}
\label{table:exp:real:coverage}
\end{scriptsize}
\vspace{-2mm}
\end{table}

\def\TabTCRUnscaled{

\begin{table}[ht]
\vspace{-3mm}
\captionof{table}{
The tail coverage rate (mean longitudinal coverage for the least-covered 10\% TS), the higher the better.
The best method is in \textbf{bold}, and the best method without using any infinitely-wide PI is \underline{underscored}.
Both versions of \methodname consistently outperform all baselines.
}
\centering
\begin{scriptsize}
\setlength\tabcolsep{6pt}
\begin{tabular}{l|ll|lll|ll}
\toprule
Tail Coverage Rate  $\uparrow$  &\methodBudget & \methodError & \baselineCFRNN & \baselineCQR & \baselineMADSplit &  \baselineQRNN  & \baselineDPRNN  \\
\midrule
\dataMIMIC & \underline{71.59$\pm$4.03} & \textbf{80.68$\pm$1.74} & 62.22$\pm$7.09 & 68.60$\pm$3.84 & 65.05$\pm$6.12 & 61.80$\pm$3.91 & 17.24$\pm$5.38\\
\dataCLAIM & \underline{74.16$\pm$1.22} & \textbf{81.53$\pm$0.77} & 65.95$\pm$1.88 & 66.45$\pm$3.19 & 68.08$\pm$2.44 & 53.89$\pm$3.59 & 1.65$\pm$0.54\\
\dataCOVID & \underline{70.01$\pm$4.45} & \textbf{82.39$\pm$1.28} & 64.41$\pm$6.11 & 66.41$\pm$5.99 & 67.38$\pm$4.63 & 65.16$\pm$6.15 & 36.65$\pm$5.63\\
\dataEEG     & \underline{70.99$\pm$2.18} & \textbf{79.03$\pm$1.22} & 64.14$\pm$3.42 & 61.95$\pm$4.71 & 67.13$\pm$2.32 & 57.82$\pm$2.78 & 12.99$\pm$1.32\\
\dataGEFCom   & \underline{68.96$\pm$1.70} & \textbf{81.77$\pm$0.36} & 58.49$\pm$1.38 & 61.63$\pm$1.56 & 60.46$\pm$1.66 & 47.56$\pm$2.27 & 67.45$\pm$1.69\\
\dataGEFComR & \underline{75.28$\pm$1.28} & \textbf{81.80$\pm$0.69} & 68.76$\pm$2.18 & 71.95$\pm$1.66 & 70.79$\pm$2.12 & 64.99$\pm$1.92 & 71.86$\pm$1.75\\
\bottomrule
\end{tabular}
\label{table:exp:real:tcr}
\end{scriptsize}
\vspace{-2mm}
\end{table}
}
\TabTCRUnscaled

\def\TabInvEff{

\begin{table}[ht]
\vspace{-3mm}
\captionof{table}{
Inverse Efficiency, measured by the mean PI width divided by the coverage rate.
Since \methodError can create infinite PI, the width is computed by replacing $\infty$ with 2x the maximum finite PI width.
The most efficient (and valid) method is in \textbf{bold} (p-value=0.01). 
As we can see, \methodBudget is highly competitive in efficiency.
}
\centering
\begin{scriptsize}
\setlength\tabcolsep{6pt}
\begin{tabular}{l|ll|lll|ll}
\toprule
Inverse Efficiency $\downarrow$    &\methodBudget & \methodError & \baselineCFRNN & \baselineCQR & \baselineMADSplit &  \baselineQRNN  & \baselineDPRNN  \\
\midrule
\dataMIMIC & 1.990$\pm$0.165 & 2.382$\pm$0.265 & 1.964$\pm$0.170 & \textbf{1.738$\pm$0.145} & 2.072$\pm$0.223 & 1.623$\pm$0.146 & 1.258$\pm$0.132\\
\dataCLAIM & 3.020$\pm$0.045 & 3.279$\pm$0.074 & 3.003$\pm$0.052 & \textbf{2.902$\pm$0.044} & 3.009$\pm$0.064 & 2.691$\pm$0.035 & 2.401$\pm$0.205\\
\dataCOVID & \textbf{0.831$\pm$0.032} & 1.167$\pm$0.337 & \textbf{0.826$\pm$0.034} & 0.908$\pm$0.091 & \textbf{0.826$\pm$0.037} & 0.888$\pm$0.096 & 0.744$\pm$0.050\\
\dataEEG & \textbf{1.449$\pm$0.025} & 1.749$\pm$0.125 & \textbf{1.445$\pm$0.031} & 1.586$\pm$0.052 & \textbf{1.448$\pm$0.025} & 1.497$\pm$0.042 & 1.061$\pm$0.027\\
\dataGEFCom & 0.238$\pm$0.005 & 0.280$\pm$0.013 & \textbf{0.235$\pm$0.005} & 0.242$\pm$0.005 & 0.238$\pm$0.005 & 0.211$\pm$0.005 & 0.636$\pm$0.009\\
\dataGEFComR & \textbf{0.200$\pm$0.004} & 0.222$\pm$0.010 & \textbf{0.198$\pm$0.004} & 0.207$\pm$0.004 & 0.201$\pm$0.004 & 0.193$\pm$0.004 & 0.590$\pm$0.009\\
\bottomrule
\end{tabular}
\label{table:exp:real:inveff}
\end{scriptsize}
\vspace{-2mm}
\end{table}
}
\TabInvEff



\section{Conclusions}
In this paper, we proposed Temporal Quantile Adjustment, or \methodname,  to quantify uncertainty (create prediction intervals) in time series forecasting with a cross-section. 
\methodname belongs to the framework of conformal prediction, and the main idea is to adjust the quantile to query using temporal information collected so far.
This allows \methodname to work with \textit{any model and any nonconformity score design}.
\methodname theoretically is ``no-worse'' in cross-sectional coverage than vanilla split conformal as long as the expected value of adjustment is zero, and empirically \textit{improves} the coverage.
We also proposed two variants, \methodBudget and \methodError, both of which significantly outperform baselines in improving temporal/longitudinal coverage across many real world datasets. We hope that this work will serve as a foundation for the future design of PIs with both high cross-sectional and temporal coverage.

\bibliography{main}
\bibliographystyle{plain}

\newpage
\section*{Checklist}

The checklist follows the references.  Please
read the checklist guidelines carefully for information on how to answer these
questions.  For each question, change the default \answerTODO{} to \answerYes{},
\answerNo{}, or \answerNA{}.  You are strongly encouraged to include a {\bf
justification to your answer}, either by referencing the appropriate section of
your paper or providing a brief inline description.  For example:
\begin{itemize}
  \item Did you include the license to the code and datasets? \answerYes{See Section...}
  \item Did you include the license to the code and datasets? \answerNo{The code and the data are proprietary.}
  \item Did you include the license to the code and datasets? \answerNA{}
\end{itemize}
Please do not modify the questions and only use the provided macros for your
answers.  Note that the Checklist section does not count towards the page
limit.  In your paper, please delete this instructions block and only keep the
Checklist section heading above along with the questions/answers below.

\begin{enumerate}

\item For all authors...
\begin{enumerate}
  \item Do the main claims made in the abstract and introduction accurately reflect the paper's contributions and scope?
    \answerYes{}
  \item Did you describe the limitations of your work?
    \answerYes{\methodname could theoretically lead to worse (although empirically better) coverage, as discussed in Section~\ref{sec:method}.}
  \item Did you discuss any potential negative societal impacts of your work?
    \answerNA{We do not foresee any potential negative societal impact of this work.}
  \item Have you read the ethics review guidelines and ensured that your paper conforms to them?
    \answerYes{}
\end{enumerate}

\item If you are including theoretical results...
\begin{enumerate}
  \item Did you state the full set of assumptions of all theoretical results?
    \answerYes{Apart from exchangeability of time series (which we assume throughout this paper), any assumption is stated right before the relevant theorem.}
        \item Did you include complete proofs of all theoretical results?
    \answerYes{In the Appendix.}
\end{enumerate}

\item If you ran experiments...
\begin{enumerate}
  \item Did you include the code, data, and instructions needed to reproduce the main experimental results (either in the supplemental material or as a URL)?
    \answerYes{In the supplemental material and will be published if accepted.}
  \item Did you specify all the training details (e.g., data splits, hyperparameters, how they were chosen)?
    \answerYes{In the Appendix.}
        \item Did you report error bars (e.g., with respect to the random seed after running experiments multiple times)?
    \answerYes{We repeat all experiments 50 times.}
        \item Did you include the total amount of compute and the type of resources used (e.g., type of GPUs, internal cluster, or cloud provider)?
    \answerYes{In Appendix.}
\end{enumerate}

\item If you are using existing assets (e.g., code, data, models) or curating/releasing new assets...
\begin{enumerate}
  \item If your work uses existing assets, did you cite the creators?
    \answerYes{}
  \item Did you mention the license of the assets?
    \answerYes{In Appendix.}
  \item Did you include any new assets either in the supplemental material or as a URL?
    \answerNA{}
  \item Did you discuss whether and how consent was obtained from people whose data you're using/curating?
    \answerYes{Most datasets are publicly available, except for \dataCLAIM which is proprietary and provided to us by a company.}
  \item Did you discuss whether the data you are using/curating contains personally identifiable information or offensive content?
    \answerYes{They have been de-identified if applicable.}
\end{enumerate}

\item If you used crowdsourcing or conducted research with human subjects...
\begin{enumerate}
  \item Did you include the full text of instructions given to participants and screenshots, if applicable?
    \answerNA{}
  \item Did you describe any potential participant risks, with links to Institutional Review Board (IRB) approvals, if applicable?
    \answerNA{}
  \item Did you include the estimated hourly wage paid to participants and the total amount spent on participant compensation?
    \answerNA{}
\end{enumerate}

\end{enumerate}

\newpage

\appendix
\section{Proofs}

\subsection{Proof for Theorem~\ref{thm:a_t:noworse}}
\begin{proof}
Denote $\hat{r}_{N+1, t} = 1-\alpha + \hat{\delta}_{N+1, t}$. 
To simplify the notation, in the following, all $\mathbb{P}$ means $\mathbb{P}_{\TS_{N+1}\sim\mathcal{P}_S}$.
We have:
\begin{align}
    \mathbb{P}\{Y_{N+1, t} \in \hat{C}_{a_{N+1,t}, N+1, t}\} &\geq \mathbb{P}\{rank_{N+1,t} \leq \hat{r}_{N+1,t}\} \\
    &=\int_0^1 \int_{0}^{q} f_{rank_{N+1,t},\hat{r}_{N+1,t}}(r, q) dr dq\\
    &= \int_0^1 \int_{0}^{q} f_{rank_{N+1,t}}(r)f_{\hat{r}_{N+1,t}}(q) dr dq\\
    &= \int_0^1 q f_{\hat{r}_{N+1,t}}(q) dq\\
    &= \mathbb{E}[\hat{r}_{N+1,t}] = 1 - \alpha + \mathbb{E}[\hat{\delta}_{N+1,t}]
\end{align}
\end{proof}

\subsection{Proof for Theorem~\ref{thm:budget:expectation}}
\begin{proof}
For any $t$, we know that $\hat{r}_{N+1, t}$ follows a uniform distribution among $\{\frac{j}{N}\}_{j=0}^{N}$ by exchangeability. 
For simplicity, we drop the subscripts $\cdot_{N+1,t}$. 
As a result, we have
\begin{align}
    \mathbb{E}[\hat{r}| \hat{r} \geq 1-\alpha] &= \frac{\ceil{(1-\alpha)N} + N}{2 N }\\
    \mathbb{E}[\hat{r}| \hat{r} < 1-\alpha] &= \frac{\ceil{(1-\alpha)N}-1}{2 N }
\end{align}
More over, 
\begin{align}
    \mathbb{P}\{\hat{r} \geq 1-\alpha\} &= 1 - \frac{\ceil{(1-\alpha)N}}{N+1}\\
    \mathbb{P}\{\hat{r} < 1-\alpha\} &= \frac{\ceil{(1-\alpha)N}}{N+1}.
\end{align}
Denoting $\epsilon\defeq \alpha N - \floor{\alpha N} = \ceil{(1-\alpha)N} - (1-\alpha)N$, this means
\begin{align}
    \mathbb{E}[g^{\methodBudgetSuffix}(\hat{r};\alpha)] &= \Bigg(1 - \frac{\ceil{(1-\alpha)N}}{N+1}\Bigg) \Bigg(\frac{\ceil{(1-\alpha)N} + N}{2 N } - (1-\alpha)\Bigg) \\ & \mspace{150mu} + C \frac{\ceil{(1-\alpha)N}}{N+1} \Bigg(\frac{\ceil{(1-\alpha)N}-1}{2 N } - (1-\alpha)\Bigg)\\
    &= \Bigg(1-\frac{(1-\alpha)N + \epsilon}{N+1}\Bigg)\Bigg(\frac{(2-\alpha)N + \epsilon}{2N} - (1-\alpha)\Bigg) \\ & \mspace{150mu} + C\frac{(1-\alpha)N + \epsilon}{N+1}\Bigg(\frac{(1-\alpha)N - 1 + \epsilon}{2N} - (1-\alpha)\Bigg)\\
    &= \Bigg(\frac{\alpha N + 1 - \epsilon}{N+1}\Bigg)\Bigg(\frac{\alpha N + \epsilon}{2N}\Bigg) \\ & \mspace{150mu} + C\frac{(1-\alpha)N + \epsilon}{N+1}\Bigg(\frac{(\alpha - 1)N - 1 + \epsilon}{2N}\Bigg)\\
    &= \frac{\alpha^2 N^2 + \alpha N + \epsilon - \epsilon^2}{2(N+1)N}  - C \frac{(1-\alpha)^2 N^2 + (1-\alpha)N + \epsilon - \epsilon^2}{2(N+1)N}\\
\end{align}
As a result,
\begin{align}
    \mathbb{E}[g^{\methodBudgetSuffix}(\hat{r};\alpha)] = 0 \Leftrightarrow C = \frac{(\alpha N + \epsilon)(\alpha N + 1 - \epsilon)}{((1-\alpha) N + \epsilon)((1-\alpha) N + 1 - \epsilon)}
\end{align}
\end{proof}

\subsection{Proof for Theorem~\ref{thm:budget:guarantee}}
\begin{proof}
The inequality trivially holds for $\alpha\geq 0.5$, so we only focus on the case when $\alpha < 0.5$.

An upper-bound for the additional loss is simply $L\defeq -\min_{q} g^{\methodBudgetSuffix}(q;\alpha)$.
To find an upper bound for $L$, notice that
\begin{align}
    L = -\min_{q} g^{\methodBudgetSuffix}(q;\alpha) &= (1-\alpha) C \\
    &= \frac{(\alpha N + \epsilon)(\alpha N + 1 - \epsilon)}{((1-\alpha) N + \epsilon)((1-\alpha) N + 1 - \epsilon)} (1-\alpha)\\
    &= \frac{\alpha^2 N^2 + \alpha N + \epsilon - \epsilon^2}{(1-\alpha)^2 N^2 + (1-\alpha) N + \epsilon - \epsilon^2} (1-\alpha)\\
    &=\Big (1 - \frac{N^2[(1-\alpha)^2 - \alpha^2] + N (1-2\alpha)}{(1-\alpha)^2 N^2 + (1-\alpha) N + \epsilon - \epsilon^2} \Big) (1-\alpha)
\end{align}
If we consider the last expression as a function of $\epsilon$, it is maximized with $\epsilon = \frac{1}{2}$, which means 
\begin{align}
    L \leq \Bigg(\frac{\alpha + \frac{1}{2N}}{1-\alpha  + \frac{1}{2N}}\Bigg)^2 (1-\alpha)
\end{align}
Denote $\alpha^+ \defeq \alpha + \Bigg(\frac{\alpha + \frac{1}{2N}}{1-\alpha  + \frac{1}{2N}}\Bigg)^2 (1-\alpha)$.
We have $\hat{C}^{split}_{\alpha^+,N+1,t+1} \subseteq \hat{C}^{\methodBudget}_{\alpha, N+1,t+1}$.
This means
\begin{align}
    \mathbb{P}_{\TS_{N+1}\sim\mathcal{P}_S}\{ Y_{N+1,t+1}\in\hat{C}^{\methodBudget}_{\alpha, N+1,t+1} \} \\
    &\geq \mathbb{P}_{\TS_{N+1}\sim\mathcal{P}_S}\{ Y_{N+1,t+1}\in \hat{C}^{split}_{\alpha^+,N+1,t+1} \} \\
    &\geq  1 - \alpha - \Bigg(\frac{\alpha + \frac{1}{2N}}{1-\alpha  + \frac{1}{2N}}\Bigg)^2 (1-\alpha)
\end{align}
\end{proof}

\subsection{Proof for Theorem~\ref{thm:aci:finite}}
\begin{proof}
First, let us consider the alternative quantile adjustment $b^{\methodError}_{N+1,t}$ with update rule:
\begin{align}
    b_{t+1} \gets  \begin{cases}
    b_t + \gamma(\alpha - err_t) & (b_t \in [0,1]) \\ 
    b_t + \gamma(\alpha - b_t) & (otherwise)
    \end{cases}\label{eq:a_t:update:alt}.
\end{align}

We will first show $\forall t, \mathbb{E}[b_t] = \alpha$ by induction, as we hinted in the main paper.
(For simplicity in notation, we drop the subscript $\cdot_{N+1}$.)
To begin with, we have 
\begin{align}
    \mathbb{E}_{\TS}[b^{\methodError}_{0}] = \mathbb{E}_{\TS}[\alpha]= \alpha
\end{align}
Now, 
\begin{align}
    \mathbb{E}_{\TS}[b_{t+1} | b_t] &= \begin{cases}
    b_t + \gamma(\alpha - \mathbb{E}_{\TS}[err_t]) & (b_t \in [0,1]) \\ 
    b_t + \gamma(\alpha - b_t) & (otherwise)
    \end{cases}
\end{align}
If the nonconformity score's rank has no temporal dependence, we have $err_t\sim Bernoulli(a_t)$ (here again we assume we use random coin-flips to ensure an exact coverage probability of $1-\alpha$), which means $\mathbb{E}[err_t] = b_t$.
This leads to
\begin{align}
    & \mathbb{E}_{\TS}[b_{t+1} | b_t] = b_t + \gamma(\alpha - b_t)\\
    \implies &\mathbb{E}_{\TS}[b_{t+1}] = \mathbb{E}_{\TS}[\mathbb{E}_{\TS}[b_{t+1} | b_t]] = (1-\gamma) \mathbb{E}_{\TS}[b_t] + \gamma \alpha = \alpha
\end{align}
By induction, we have 
\begin{align}
    \forall t, \mathbb{E}_{\TS}[b_t] = \alpha
\end{align}

Next, we should compare $\mathbb{E}[a_t]$ and $\mathbb{E}[b_t]$.
It should be clear that 
\begin{align}
    \forall c, \mathbb{E}_{\TS}[a_{t+1} | a_t = c] \leq \mathbb{E}_{\TS}[b_{t+1} | b_t = c].
\end{align}
To begin with, we have $\mathbb{E}[a_0] \leq \alpha$. 
Following a similar logic, we have
\begin{align}
    \mathbb{E}_{\TS}[a_{t+1}] = \mathbb{E}_{\TS}[\mathbb{E}_{\TS}[a_{t+1} | a_t]] \leq (1-\gamma) \mathbb{E}_{\TS}[a_t] + \gamma \alpha \leq \alpha
\end{align}
By induction, we get
\begin{align}
    \forall t, \mathbb{E}_{\TS_{N+1}\sim\mathcal{P}_S}[a^{\methodError}_{N+1,t}] \leq \alpha.
\end{align}

\end{proof}
\textbf{Remarks}
The alternative update rule in Eq.~\ref{eq:a_t:update:alt} is not just used to prove~Theorem~\ref{thm:aci:finite}. 
In fact, one might want to use this rule in practice if the TS is not very long and the asymptotic guarantee (Theorem~\ref{thm:aci:asymp}) is not relevant. 
This is because Eq.~\ref{eq:a_t:update:alt} does not become more conservative on average as $t$ increases. 

\subsection{Proof for Theorem~\ref{thm:aci:asymp}}
\begin{proof}
We will use an similar argument similar to one in~\cite{ACI}.
First, we have $a_t \in [-\gamma, 1+\gamma]$ if $a_0 = \alpha$.
To see this, note that $a_{t+1} > a_t \implies a_t \leq 1 \implies a_{t+1} \leq 1+\gamma$, which means $\forall t, a_t \leq 1+\gamma$. 
Similarly, $a_{t+1} < a_t \implies a_t \geq 0 \implies a_{t+1} \geq -\gamma$, which means $\forall t, a_t \geq -\gamma$. 

Note that with Eq.~\ref{eq:a_t:update}, we have $a_{t+1} \leq a_t + \gamma (\alpha - a_t)$.
Now, if we expand $a_T$, we get
\begin{align}
    & -\gamma \leq a_T \leq \alpha + \sum_{t=0}^{T-1} \gamma (\alpha - err_t) \\
    \implies & \frac{\sum_{t=0}^{T-1} err_t}{T} \leq \alpha + \frac{\alpha + \gamma}{\gamma T}
\end{align}
By taking the limit of both sides, we are done.

\end{proof}

\newpage


\newpage
\section{Additional Experiment Details}

\def\TabWidthMedian{

\begin{table}[ht]
\captionof{table}{
Median of PI width.
The most efficient (and valid) method is in \textbf{bold}.
}
\centering
\begin{scriptsize}
\setlength\tabcolsep{6pt}
\begin{tabular}{l|cc|ccc|cr}
\toprule
Width    &\methodBudget & \methodError & \baselineCFRNN & \baselineCQR & \baselineMADSplit &  \baselineQRNN  & \baselineDPRNN  \\
\midrule
\dataMIMIC & 1.672$\pm$0.163 & 1.639$\pm$0.146 & 1.753$\pm$0.169 & \textbf{1.443$\pm$0.120} & 1.640$\pm$0.172 & 1.279$\pm$0.116 & 0.493$\pm$0.023\\
\dataCLAIM & 2.642$\pm$0.052 & 2.561$\pm$0.051 & 2.710$\pm$0.060 & 2.628$\pm$0.048 & \textbf{2.456$\pm$0.057} & 2.325$\pm$0.032 & 0.434$\pm$0.023\\
\dataCOVID    & 0.705$\pm$0.033 & \textbf{0.696$\pm$0.033} & 0.727$\pm$0.037 & 0.767$\pm$0.069 & 0.699$\pm$0.031 & 0.748$\pm$0.082 & 0.386$\pm$0.052\\
\dataEEG     & 6.635$\pm$0.095 & \textbf{5.336$\pm$0.058} & 6.742$\pm$0.098 & 6.776$\pm$0.102 & 6.743$\pm$0.098 & 6.805$\pm$0.134 & 0.100$\pm$0.024\\
\dataGEFCom   & 0.198$\pm$0.004 & 0.196$\pm$0.004 & 0.198$\pm$0.004 & 0.204$\pm$0.004 & \textbf{0.191$\pm$0.004} & 0.155$\pm$0.005 & 0.428$\pm$0.006\\
\dataGEFComR & 0.169$\pm$0.008 & 0.174$\pm$0.005 & 0.170$\pm$0.008 & 0.171$\pm$0.004 & \textbf{0.164$\pm$0.005} & 0.149$\pm$0.005 & 0.458$\pm$0.009\\
\bottomrule
\end{tabular}
\label{table:exp:real:width:median}
\end{scriptsize}
\end{table}
}
\def\TabWidthMean{

\begin{table}[ht]
\captionof{table}{
Mean PI Width.
Since \methodError can create infinite PI, the mean is computed by replacing $\infty$ with 2x the maximum finite PI width.
The most efficient (and valid) method is in \textbf{bold}.
As we can see, CFRNN tends to have the narrowest PI, but all methods (except for \methodError due to the adjustment for infinite PIs) have very similar mean widths.
}
\centering
\begin{scriptsize}
\setlength\tabcolsep{6pt}
\begin{tabular}{l|ll|lll|ll}
\toprule
Width   $\downarrow$   &\methodBudget & \methodError & \baselineCFRNN & \baselineCQR & \baselineMADSplit &  \baselineQRNN  & \baselineDPRNN  \\
\midrule
\dataMIMIC & 1.760$\pm$0.153 & 2.189$\pm$0.245 & 1.769$\pm$0.164 & \textbf{1.566$\pm$0.132} & 1.872$\pm$0.212 & 1.411$\pm$0.130 & 0.578$\pm$0.031\\
\dataCLAIM & 2.743$\pm$0.050 & 3.002$\pm$0.070 & \textbf{2.709$\pm$0.056} & 2.616$\pm$0.049 & 2.714$\pm$0.067 & 2.312$\pm$0.033 & 0.594$\pm$0.042\\
\dataCOVID    & \textbf{0.725$\pm$0.029} & 1.056$\pm$0.307 & 0.735$\pm$0.035 & 0.806$\pm$0.081 & 0.736$\pm$0.037 & 0.782$\pm$0.085 & 0.498$\pm$0.049\\
\dataEEG     & 6.542$\pm$0.067 & 6.553$\pm$0.083 & \textbf{6.396$\pm$0.079} & 6.437$\pm$0.082 & 6.397$\pm$0.079 & 6.420$\pm$0.105 & 0.171$\pm$0.022\\
\dataGEFCom   & 0.211$\pm$0.004 & 0.255$\pm$0.012 & \textbf{0.208$\pm$0.004} & 0.216$\pm$0.004 & 0.212$\pm$0.005 & 0.170$\pm$0.005 & 0.569$\pm$0.007\\
\dataGEFComR & 0.179$\pm$0.004 & 0.201$\pm$0.010 & \textbf{0.178$\pm$0.005} & 0.187$\pm$0.004 & 0.181$\pm$0.005 & 0.165$\pm$0.005 & 0.537$\pm$0.011\\
\bottomrule
\end{tabular}
\label{table:exp:real:width:mean}
\end{scriptsize}
\end{table}
}

\subsection{Datasets}
\textbf{\dataMIMIC}~\cite{MIMIC,PhysioNet,MIMICDemo}: 
MIMIC-III is a large public (but requires application of access) database consisting of records for patients admitted to critical care units. 
The task is to predict white blood cell counts (WBCC) levels, using 25 features (following~\cite{alaaBlockRNN}) including measurements like systolic/diastolic blood pressure or the dosage of antibiotics.
To prepare the data, we follow~\cite{alaaCFRNN} and restrict our study to patients on antibiotics (Levofloxacin).
We drop sequences with less than 30 visits, ending up with 392 sequences in total.
The exact SQL queries and data processing scripts are provided in the supplemental material and will be published. 
Note that the EHR has been de-identified - in particular, all visits for a patient are shifted in dates, so we only know the relative dates of visits belonging to the same patient, but do not know the actual dates, making it impossible to perform any temporal splitting.

\textbf{\dataCLAIM}: 
\dataCLAIM is a proprietary dataset from a large healthcare data provider in North America. 
It contains longitudinal view of information like inpatient and outpatient services, prescription, costs or enrollment.
The task is to predict the (allowed) claim amount for the next visit.
The features include diagnoses codes (ICD-10) and procedure codes (CPT/HCPCS), the gender and age of patient, and the previous claim amount.

\textbf{\dataCOVID}~\cite{COVID}: 
\dataCOVID is available at \url{https://coronavirus.data.gov.uk/}. 
We follow the setup of~\cite{alaaCFRNN} and treat different regions within UK as the cross-section, and we used the most recent full-month data available at the time of this work (March 2022).
The feature only consists of previous day's case number.
We include \dataCOVID for completeness as it was used in~\cite{alaaCFRNN}, and as an idealized experiment, despite its unrealistic settings. 

\textbf{\dataEEG}~\cite{UCI_EEG}:
\dataEEG is available at \url{https://archive.ics.uci.edu/ml/datasets/EEG+Database}. 
The ``Large'' version of the dataset contains the electroencephalography (EEG) signals sampled at 256 Hz for 1 second for 10 alcoholic and 10 control subjects.
Like~\cite{alaaCFRNN}, we used only the control group, and downsampled the signals to facilitate training, but to length 64 instead of 50 to avoid unnecessary interpolation. 
We do not mix data from different EEG channels, and use only channel 0 data because different EEG channels should follow different developments.
The feature only consists of the EEG signal at the previous step.

\textbf{\dataGEFCom}~\cite{GEFCom}:
\dataGEFCom (2014) is the Probabilistic Electric Load Forecasting task in Global Energy Forecasting Competition 2014.
It has hourly temperature and electricity load data for one utility. 
The load data covers 9 years.
For this task, we split the data into 24-hour-sequences, and consider these sequences as forming a cross-section. 
The features consist of temperature data from 25 stations, and the previous load.

\textbf{Data Licenses and Consent}:
\begin{itemize}
    \item \dataMIMIC: The PhysioNet Credentialed Health Data License. The data is public, but requires access application.
    \item \dataCOVID: Open Government Licence v3.0.
    Publicly available on \url{https://coronavirus.data.gov.uk/}.
    \item \dataCLAIM: This is proprietary data.
    \item \dataEEG: There are no usage restrictions on this data per \url{https://archive.ics.uci.edu/ml/datasets/eeg+database}.
    \item \dataGEFCom: We could not find the license, but it is publicly available (provided by the author) on \url{http://blog.drhongtao.com/2017/03/gefcom2014-load-forecasting-data.html}.
\end{itemize}

\subsection{Implementation and Experiment Details}

We use mostly the same architecture and training protocols as~\cite{alaaCFRNN}.
The RNN (mean estimator) is a one-layer LSTM~\cite{LSTM}, with an embedding size of 32. 
The optimizer is ADAM~\cite{Adam}, and the learning rate is 1e-3. 
The training epochs for \dataMIMIC/\dataCLAIM/\dataCOVID/\dataEEG/\dataGEFCom are 200/500/1000/100/1000.
We use the same settings for the residual predictor (for \baselineMADSplit) and \baselineQRNN, except that 
\begin{itemize}
    \item The target for the residual predictor is replaced by $|y-\hat{y}|$.
    \item The loss for \baselineQRNN is replaced with the quantile loss, and the output is two scalars instead of one. 
\end{itemize}

We normalize each feature and $Y$ by the mean and standard deviation in order for the underlying LSTM to train. 
In each repetition of a experiments, we pool all data together and re-split them into training/calibration/test set using a new seed, and the LSTMs are also trained using different seeds each time.
The only exception is \dataGEFCom (the non-R version), which has only one data splitting due to the temporal order.
The LSTMs for \dataGEFCom are still trained using different seeds.
All code is provided in the supplemental material and will be published after the review period.Experiments in this paper are fast to run on a personal laptop - the largest single experiment is for \dataCLAIM (without repetition), which takes about half an hour, and all experiments finish within 2 hours (for one seed).

\section{Variants of \methodBudget}

\subsection{Rank-based Quantile Prediction}
Instead of the quantile prediction we used in the paper, we also tried the following prediction. 
$\hat{r}^{ewa}_{i,t+1} = Q^{-1}(\overline{r}_{i,t}; \{\overline{r}_{j,t}\}_{j=1}^{N+1})$ where $\overline{r}_{i,t} \defeq \frac{\sum_{t'=1}^{t} \beta^{t+1-t'} r_{i,t'}}{\sum_{t'=1}^{t}  \beta^{t+1-t'}}$.
This is an exponentially weighted (decaying) average of the past ranks.
We use $\beta=0.8$, and refer to such quantile prediction as \textbf{Rank}-based (as opposed to \textbf{Scaled}-based in Section~\ref{sec:method:budget}).

\subsection{The Aggressive Quantile Budgeting}

\def \AppendixFigTQAAlt{
\begin{figure}[ht]
    \centering
    \includegraphics[width=1\textwidth]{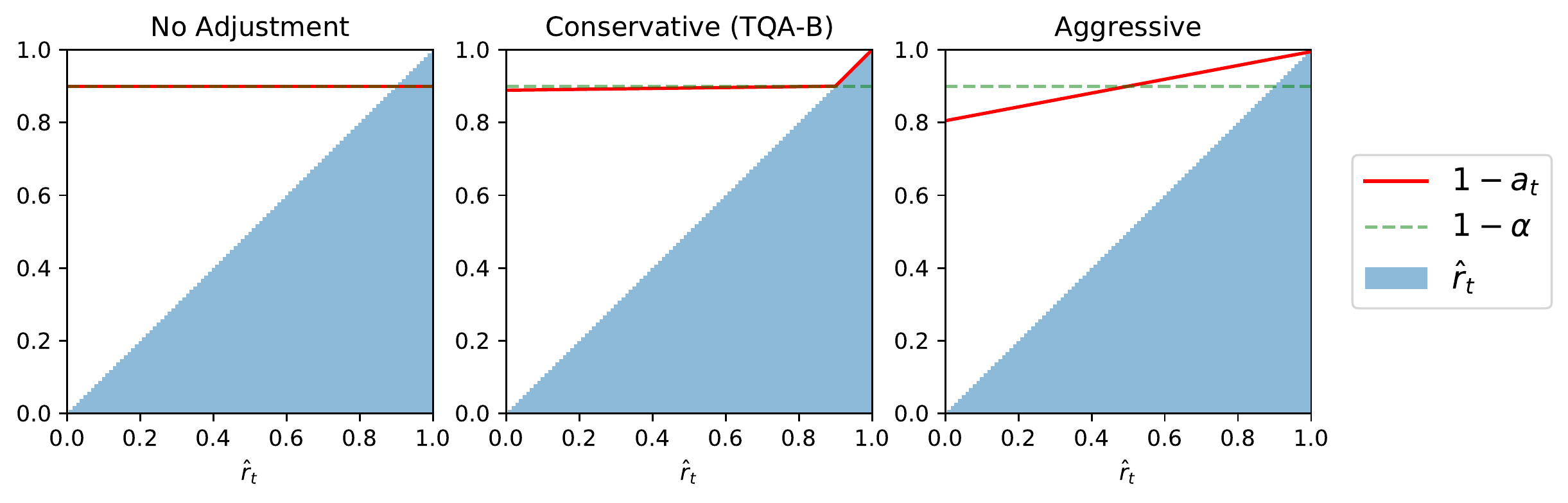}
\caption{
Here we present visualizations of different quantile budgeting methods.
``Budgeting'' refers to the fact that all red lines are on average $1-\alpha$.
The first picture (No Adjustment) is the same as the basic split conformal method.
The second picture refers to the quantile budgeting method presented in Section~\ref{sec:method:budget}.
It is conservative because even for a very small $\hat{r}_t$ like 0, the adjusted $a_t$ is still close to $\alpha$.
The third picture is the more aggressive budgeting method, and it could set $a_t$ to as big as $2\alpha$.
In a sense, it trusts the prediction $\hat{r}_t$ and adapts to it more.
As a result, if $\hat{r}_t$ predicts the actual rank of nonconformity scores well, it can achieve a much higher coverage.
The extreme case is if $\hat{r}_t$ perfectly predicts the realized ranks: ``No Adjustment'' will still have $1-\alpha$ coverage whereas the other two will achieve perfect coverage. 
\label{appendix:fig:tqa_alt}
}
\end{figure}
}
\AppendixFigTQAAlt

In this section, we present an alternative quantile budgeting function to $g^{\methodBudgetSuffix}$.
The specific adjustment is given by 
\begin{align}
    g^{\methodCenteredSuffix}(q;\alpha) &\defeq \frac{q-0.5}{2\alpha}.
\end{align}
Like in Section~\ref{sec:method:budget}, one could use a $\lambda$ to ensure the final $a_t$ is not too close to zero (to avoid infinitely-wide PIs).
We refer to this ``aggressive'' version as \methodCentered.
Figure~\ref{appendix:fig:tqa_alt} visualizes the difference between ``conservative'' (\methodBudget) and ``aggressive'' budgeting. 
It should be clear that \methodCentered also satisfies Theorem~\ref{thm:budget:expectation},
Theorem~\ref{thm:budget:guarantee}, however, becomes looser:
\begin{theorem}
\begin{align}
\mathbb{P}_{\TS_{N+1}\sim\mathcal{P}_S}\{ Y_{N+1,t+1}\in\hat{C}^{\methodCentered}_{\alpha, N+1,t+1} \} \geq 1-2\alpha.
\end{align}
\end{theorem}
The proof is essentially the same as that for Theorem~\ref{thm:budget:guarantee}.

\subsection{Comparison}
The full results of different variants of \methodBudget are in Table~\ref{table:exp:appendix:variants}.
All variants are generally valid (except for a tiny drop in coverage for Rank+Conservative on \dataGEFComR).
In general, among the two quantile prediction methods, Scale tends to produce better tail coverage rate, at the cost of worse efficiency. 
Among the quantile budgeting methods, the Aggressive version also tends to increase tail coverage rate and decrease efficiency. 
In practice, one might design the two components depending on the whether efficiency or tail coverage rate is more important.

\def\TabAppendixVariants{

\begin{table}[ht]
\captionof{table}{
Mean coverage, tail coverage, and inverse efficiency for different variants of \methodBudget.
Valid mean coverage and the best of tail coverage and inverse efficiency are in \textbf{bold}.
Note that we re-scaled all methods' mean PI width to the same as CFRNN for the TCR here, for a fair comparison. 
(We couldn't do this in a meaningful way for \methodError in the main text.)
}
\centering
\begin{scriptsize}
\setlength\tabcolsep{6pt}
\begin{tabular}{l|ll|ll|l}
\toprule
 & \multicolumn{4}{c|}{\methodBudget} & \baselineCFRNN\\
\midrule
Quantile Prediction Method & \multicolumn{2}{c|}{Scale} & \multicolumn{2}{c|}{Rank} & \\
\midrule
 Budgeting Method & \underline{Conservative} & Aggressive & Conservative &  Aggressive & \\
\midrule
Coverage    &\\
\midrule
\dataMIMIC & \textbf{91.31$\pm$1.32} & \textbf{93.62$\pm$1.04} & \textbf{89.79$\pm$1.68} & \textbf{91.79$\pm$1.41} & \textbf{90.06$\pm$1.73}\\
\dataCLAIM & \textbf{91.19$\pm$0.48} & \textbf{92.87$\pm$0.37} & \textbf{89.88$\pm$0.55} & \textbf{91.54$\pm$0.46} & \textbf{90.21$\pm$0.56}\\
\dataCOVID & \textbf{90.79$\pm$1.45} & \textbf{92.46$\pm$1.15} & \textbf{89.94$\pm$1.71} & \textbf{91.16$\pm$1.44} & \textbf{90.25$\pm$1.69}\\
\dataEEG     & \textbf{90.73$\pm$1.21} & \textbf{92.53$\pm$1.00} & \textbf{89.57$\pm$1.46} & \textbf{90.95$\pm$1.28} & \textbf{89.92$\pm$1.44}\\
\dataGEFCom   & 89.58$\pm$0.25 & \textbf{91.19$\pm$0.39} & 87.96$\pm$0.20 & 89.62$\pm$0.22 & 88.61$\pm$0.16\\
\dataGEFComR & \textbf{90.56$\pm$0.64} & \textbf{91.52$\pm$0.57} & 89.55$\pm$0.80 & \textbf{90.57$\pm$0.67} & \textbf{89.92$\pm$0.78}\\
\midrule
Tail Coverage Rate (rescaled) $\uparrow$    &\\
\midrule
\dataMIMIC & 70.18$\pm$4.17 & \textbf{76.84$\pm$4.20} & 63.06$\pm$6.46 & 70.98$\pm$4.69 & 62.22$\pm$7.09\\
\dataCLAIM & 73.36$\pm$1.27 & \textbf{78.15$\pm$1.20} & 66.73$\pm$1.76 & 73.08$\pm$1.29 & 65.95$\pm$1.88\\
\dataCOVID & 69.48$\pm$4.48 & \textbf{75.49$\pm$4.89} & 64.85$\pm$5.80 & 69.44$\pm$5.51 & 64.41$\pm$6.11\\
\dataEEG     & 70.29$\pm$2.15 & \textbf{75.81$\pm$2.69} & 64.50$\pm$3.32 & 69.51$\pm$3.01 & 64.14$\pm$3.42\\
\dataGEFCom   & 67.71$\pm$1.68 & \textbf{74.19$\pm$2.55} & 58.51$\pm$1.27 & 67.52$\pm$1.42 & 58.49$\pm$1.38\\
\dataGEFComR & 74.48$\pm$1.36 & \textbf{77.47$\pm$1.36} & 69.18$\pm$2.14 & 73.59$\pm$1.58 & 68.76$\pm$2.18\\
\midrule
Inverse Efficiency $\downarrow$    &\\
\midrule
\dataMIMIC & \textbf{1.990$\pm$0.165} & 2.128$\pm$0.182 & \textbf{1.936$\pm$0.166} & \textbf{1.979$\pm$0.166} & \textbf{1.964$\pm$0.170}\\
\dataCLAIM & 3.020$\pm$0.045 & 3.224$\pm$0.072 & \textbf{2.956$\pm$0.051} & 3.025$\pm$0.046 & 3.003$\pm$0.052\\
\dataCOVID & \textbf{0.831$\pm$0.032} & 0.858$\pm$0.036 & \textbf{0.819$\pm$0.033} & \textbf{0.828$\pm$0.033} & \textbf{0.826$\pm$0.034}\\
\dataEEG     & 1.449$\pm$0.025 & 1.498$\pm$0.023 & \textbf{1.429$\pm$0.029} & 1.444$\pm$0.025 & 1.445$\pm$0.031\\
\dataGEFCom   & 0.238$\pm$0.005 & 0.247$\pm$0.007 & \textbf{0.233$\pm$0.005} & \textbf{0.236$\pm$0.005} & \textbf{0.235$\pm$0.005}\\
\dataGEFComR & 0.200$\pm$0.004 & 0.210$\pm$0.006 & \textbf{0.196$\pm$0.004} & 0.199$\pm$0.004 & 0.198$\pm$0.004\\
\bottomrule
\end{tabular}
\label{table:exp:appendix:variants}
\end{scriptsize}
\end{table}
}
\TabAppendixVariants

\section{Additional Results}

\textbf{Full Time Series}:
We repeat the experiments in Section~\ref{sec:exp} but measure the evaluation metrics on the entire time series as opposed to the last 20 steps. 
The results are in Table~\ref{table:exp:appendix:full} and the conclusion stays the same.

\def\TabAppendixFullTS{

\begin{table}[ht]
\captionof{table}{
Mean coverage, tail coverage, and inverse efficiency using the entire time series (as opposed to the last 20 steps).
Valid mean coverage and the best of tail coverage and inverse efficiency are in \textbf{bold}.
The conclusion is the same as in the main text - \methodname greatly improves the longitudinal coverage for the least-covered TSs, and in particular \methodBudget maintains very competitive efficiency. 
}
\centering
\begin{scriptsize}
\setlength\tabcolsep{6pt}
\begin{tabular}{l|ll|lll|ll}
\toprule
Coverage     &\methodBudget & \methodError & \baselineCFRNN & \baselineCQR & \baselineMADSplit &  \baselineQRNN  & \baselineDPRNN  \\
\midrule
\dataMIMIC & \textbf{91.16$\pm$1.15} & \textbf{91.76$\pm$0.63} & \textbf{90.03$\pm$1.49} & \textbf{90.06$\pm$1.17} & \textbf{90.19$\pm$1.32} & 84.81$\pm$1.13 & 44.78$\pm$3.61\\
\dataCLAIM & \textbf{91.00$\pm$0.37} & \textbf{91.38$\pm$0.23} & \textbf{90.13$\pm$0.48} & \textbf{90.07$\pm$0.54} & \textbf{90.11$\pm$0.44} & 86.38$\pm$0.62 & 25.31$\pm$0.72\\
\dataCOVID & \textbf{90.70$\pm$1.39} & \textbf{91.73$\pm$0.74} & \textbf{90.18$\pm$1.61} & \textbf{90.08$\pm$1.44} & \textbf{90.15$\pm$1.39} & 89.11$\pm$1.45 & 65.40$\pm$2.74\\
\dataEEG & \textbf{90.56$\pm$0.85} & \textbf{91.04$\pm$0.30} & \textbf{89.82$\pm$1.07} & \textbf{89.99$\pm$1.14} & \textbf{89.82$\pm$0.96} & 86.82$\pm$0.62 & 35.35$\pm$1.18\\
\dataGEFCom & 89.34$\pm$0.23 & \textbf{90.62$\pm$0.12} & 88.50$\pm$0.15 & 89.05$\pm$0.16 & 88.83$\pm$0.16 & 80.58$\pm$1.29 & \textbf{90.46$\pm$0.73}\\
\dataGEFComR & \textbf{90.56$\pm$0.60} & \textbf{90.69$\pm$0.39} & \textbf{89.95$\pm$0.75} & \textbf{90.13$\pm$0.61} & \textbf{89.97$\pm$0.66} & 85.86$\pm$1.08 & \textbf{91.97$\pm$0.71}\\
\midrule
Tail Coverage Rate $\uparrow$    &\\
\midrule
\dataMIMIC & 75.22$\pm$3.08 & \textbf{84.64$\pm$0.95} & 67.54$\pm$4.99 & 71.42$\pm$3.58 & 69.76$\pm$4.47 & 62.53$\pm$3.73 & 21.16$\pm$4.37\\
\dataCLAIM & 76.26$\pm$0.91 & \textbf{84.60$\pm$0.39} & 68.76$\pm$1.71 & 69.75$\pm$2.50 & 71.21$\pm$1.87 & 59.78$\pm$2.85 & 4.55$\pm$0.56\\
\dataCOVID & 70.47$\pm$4.65 & \textbf{84.21$\pm$0.89} & 65.22$\pm$5.83 & 68.70$\pm$5.71 & 69.07$\pm$4.59 & 67.19$\pm$5.72 & 40.74$\pm$4.24\\
\dataEEG & 75.80$\pm$1.25 & \textbf{87.22$\pm$0.13} & 70.43$\pm$2.17 & 69.94$\pm$2.80 & 71.62$\pm$1.77 & 65.22$\pm$1.93 & 18.86$\pm$0.94\\
\dataGEFCom & 67.84$\pm$1.66 & \textbf{82.71$\pm$0.15} & 58.26$\pm$1.20 & 61.72$\pm$1.48 & 60.36$\pm$1.63 & 48.51$\pm$2.06 & 71.45$\pm$1.54\\
\dataGEFComR & 75.41$\pm$1.26 & \textbf{83.00$\pm$0.31} & 69.14$\pm$2.10 & 72.96$\pm$1.57 & 71.51$\pm$1.88 & 66.64$\pm$1.98 & 75.65$\pm$1.71\\
\midrule
Inverse Efficiency $\downarrow$    &\\
\midrule
\dataMIMIC & 2.053$\pm$0.157 & 2.478$\pm$0.333 & 2.018$\pm$0.152 & \textbf{1.831$\pm$0.153} & 2.115$\pm$0.185 & 1.676$\pm$0.148 & 1.277$\pm$0.120\\
\dataCLAIM & 3.039$\pm$0.038 & 3.245$\pm$0.061 & 3.019$\pm$0.043 & \textbf{2.934$\pm$0.039} & 3.023$\pm$0.051 & 2.740$\pm$0.031 & 2.242$\pm$0.146\\
\dataCOVID & \textbf{0.823$\pm$0.031} & 1.112$\pm$0.246 & \textbf{0.819$\pm$0.033} & 0.876$\pm$0.073 & \textbf{0.817$\pm$0.038} & 0.857$\pm$0.077 & 0.728$\pm$0.047\\
\dataEEG & \textbf{1.374$\pm$0.016} & 1.662$\pm$0.093 & \textbf{1.368$\pm$0.022} & 1.494$\pm$0.039 & \textbf{1.368$\pm$0.018} & 1.414$\pm$0.033 & 0.985$\pm$0.017\\
\dataGEFCom & 0.222$\pm$0.005 & 0.257$\pm$0.011 & \textbf{0.219$\pm$0.004} & 0.225$\pm$0.004 & 0.221$\pm$0.004 & 0.195$\pm$0.004 & 0.618$\pm$0.009\\
\dataGEFComR & \textbf{0.185$\pm$0.003} & 0.203$\pm$0.009 & \textbf{0.183$\pm$0.003} & 0.191$\pm$0.004 & 0.186$\pm$0.004 & 0.178$\pm$0.004 & 0.589$\pm$0.009\\
\bottomrule
\end{tabular}
\label{table:exp:appendix:full}
\end{scriptsize}
\end{table}
}
\TabAppendixFullTS

\textbf{Percentage of Infinitely-wide PIs}:
Since \methodError could generate infinitely-wide PIs, we check how many such intervals are actually created.
The results are in Table~\ref{table:exp:appendix:inf_pis}, and it shows that most $\hat{C}^{\methodError}$ are finite.
\def\TabAppendixPercInf{

\begin{table}[ht]
\captionof{table}{
The percentage of infinitely-wide PIs created by \methodError for different datasets. 
Only a very small percentage of the PIs are infinitely-wide.
}
\centering
\begin{scriptsize}
\setlength\tabcolsep{6pt}
\begin{tabular}{l|llllll}
\toprule
\% Infinitely-wide PI & \dataMIMIC & \dataCLAIM & \dataCOVID & \dataEEG & \dataGEFCom & \dataGEFComR\\
\midrule
Last 20 steps & 3.10$\pm$1.23 & 2.18$\pm$0.36 & 3.40$\pm$1.47 & 3.78$\pm$0.97 & 3.84$\pm$0.17 & 2.31$\pm$0.52\\
\midrule
Full TS & 2.51$\pm$0.99 & 1.72$\pm$0.28 & 2.74$\pm$1.24 & 3.46$\pm$0.78 & 3.22$\pm$0.15 & 1.94$\pm$0.44\\
\bottomrule
\end{tabular}
\label{table:exp:appendix:inf_pis}
\end{scriptsize}
\end{table}
}
\TabAppendixPercInf

\textbf{Median and Mean Width of the PIs}:
We examine the raw width of the PIs.
Note that because \methodBudget tends to improve the coverage slightly, the width is slightly higher.
In addition, the mean width should be even higher due to the attempt to cover extreme residuals (as discussed/showed in Figure~\ref{wrapfig:exp:resid}). 
However, from Table~\ref{table:exp:appendix:width}, we note that the difference in the raw PI widths are also small.

\def\TabAppendixWidth{

\begin{table}[ht]
\captionof{table}{
Median and mean width of the PIs.
The widths of infinitely-wide $\hat{C}^{\methodError}$ are replaced with 2x the width of the widest finite PI.
}
\centering
\begin{scriptsize}
\setlength\tabcolsep{6pt}
\begin{tabular}{l|ll|lll|ll}
\toprule
Median Width $\downarrow$     &\methodBudget & \methodError & \baselineCFRNN & \baselineCQR & \baselineMADSplit &  \baselineQRNN  & \baselineDPRNN  \\
\midrule
\dataMIMIC & 1.741$\pm$0.165 & 1.639$\pm$0.146 & 1.753$\pm$0.169 & 1.443$\pm$0.120 & 1.640$\pm$0.172 & 1.279$\pm$0.116 & 0.493$\pm$0.023\\
\dataCLAIM & 2.667$\pm$0.053 & 2.561$\pm$0.051 & 2.710$\pm$0.060 & 2.628$\pm$0.048 & 2.456$\pm$0.057 & 2.325$\pm$0.032 & 0.434$\pm$0.023\\
\dataCOVID & 0.735$\pm$0.036 & 0.705$\pm$0.035 & 0.739$\pm$0.038 & 0.776$\pm$0.072 & 0.706$\pm$0.034 & 0.757$\pm$0.084 & 0.390$\pm$0.052\\
\dataEEG & 1.285$\pm$0.044 & 1.207$\pm$0.035 & 1.291$\pm$0.052 & 1.343$\pm$0.069 & 1.228$\pm$0.039 & 1.218$\pm$0.040 & 0.312$\pm$0.013\\
\dataGEFCom & 0.201$\pm$0.004 & 0.196$\pm$0.004 & 0.198$\pm$0.004 & 0.204$\pm$0.004 & 0.191$\pm$0.004 & 0.155$\pm$0.005 & 0.428$\pm$0.006\\
\dataGEFComR & 0.171$\pm$0.008 & 0.174$\pm$0.005 & 0.170$\pm$0.008 & 0.171$\pm$0.004 & 0.164$\pm$0.005 & 0.149$\pm$0.005 & 0.458$\pm$0.009\\
\midrule
Mean Width $\downarrow$ & \\
\midrule
\dataMIMIC & 1.818$\pm$0.157 & 2.189$\pm$0.245 & 1.769$\pm$0.164 & 1.566$\pm$0.132 & 1.872$\pm$0.212 & 1.411$\pm$0.130 & 0.578$\pm$0.031\\
\dataCLAIM & 2.754$\pm$0.048 & 3.002$\pm$0.070 & 2.709$\pm$0.056 & 2.616$\pm$0.049 & 2.714$\pm$0.067 & 2.312$\pm$0.033 & 0.594$\pm$0.042\\
\dataCOVID & 0.755$\pm$0.033 & 1.070$\pm$0.308 & 0.746$\pm$0.038 & 0.818$\pm$0.084 & 0.745$\pm$0.037 & 0.792$\pm$0.088 & 0.503$\pm$0.050\\
\dataEEG & 1.315$\pm$0.039 & 1.585$\pm$0.111 & 1.299$\pm$0.048 & 1.428$\pm$0.069 & 1.300$\pm$0.037 & 1.317$\pm$0.043 & 0.416$\pm$0.017\\
\dataGEFCom & 0.213$\pm$0.005 & 0.255$\pm$0.012 & 0.208$\pm$0.004 & 0.216$\pm$0.004 & 0.212$\pm$0.005 & 0.170$\pm$0.005 & 0.569$\pm$0.007\\
\dataGEFComR & 0.181$\pm$0.004 & 0.201$\pm$0.010 & 0.178$\pm$0.005 & 0.187$\pm$0.004 & 0.181$\pm$0.005 & 0.165$\pm$0.005 & 0.537$\pm$0.011\\
\bottomrule
\end{tabular}
\label{table:exp:appendix:width}
\end{scriptsize}
\end{table}
}
\TabAppendixWidth

\textbf{CFRNN with Bonferroni Correction} (CFRNN-Bon):
As explained in Section~\ref{sec:prelim:scp}, to perform the correct split conformal prediction, we need to use $\infty$ in place of $v_{N+1}$ (in order to avoid plugging in different $y$ values, usually defined on a fine-granular grid, which is very expensive).
This means that, if we perform Bonferroni Correction as proposed in~\cite{alaaCFRNN} by querying the $1-\frac{\alpha}{T}$ instead of $1-\alpha$, the PI will \textit{always} by infinitely wide if $\frac{\alpha}{T} <= \frac{1}{N+1}$. 
In our experiments, this is the case for \dataMIMIC, \dataCOVID, \dataEEG and \dataGEFCom/\dataGEFComR.
The original paper~\cite{alaaCFRNN} implemented the split-conformal incorrectly by ignoring $v_{N+1}$, which is why this issue did not appear.
In addition, even if we ignore the issue of (constant) infinitely-wide PIs, in our settings the length of the time-series (e.g. number of visits by a patient) might not be known in advance.
As a result, it is not clear how to perform the Bonferroni Correction.

\textbf{Tail coverage rate over time}
Figure~\ref{fig:appendix:tcr_by_t} shows the tail coverage rate at different values of $t$.
\texttt{Ideal} means the coverage is independent for different $t$ within each time series. 
In general \methodname improves the tail coverage noticeably, but there is still gap between \texttt{Ideal} and \methodname.
\methodError sometimes approaches this optimal case (at the expense of creating infinitely-wide PIs).
\def \FigAppendixTCRByt{
\begin{figure}[ht]
    \centering
    \includegraphics[width=1\textwidth]{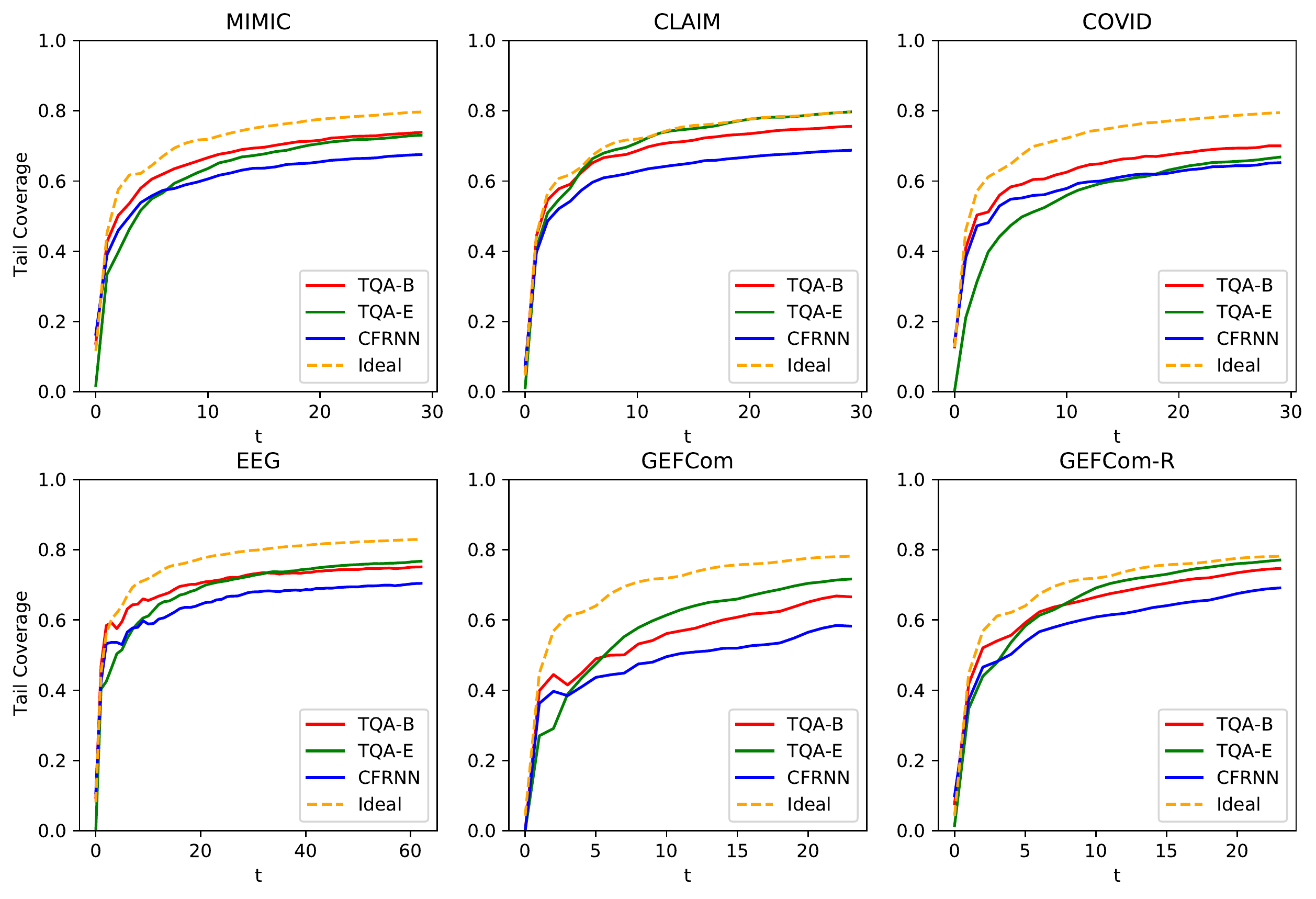}
\caption
{
Tail Coverage Rate as a function of time.
We plot the mean of 50 experiments.
There is, however, still gap between \methodname and \texttt{Ideal}.
\label{fig:appendix:tcr_by_t}
}
\end{figure}
}
\FigAppendixTCRByt

\textbf{Mean (cross-sectional) coverage over time}
Figure~\ref{fig:appendix:cov_by_t} shows the mean coverage rate for different $t$.
As expected, conformal methods show cross-sectional validity, whereas the coverage rate for non-conformal methods vary.
\def \FigAppendixCovByt{
\begin{figure}[ht]
    \centering
    \includegraphics[width=1\textwidth]{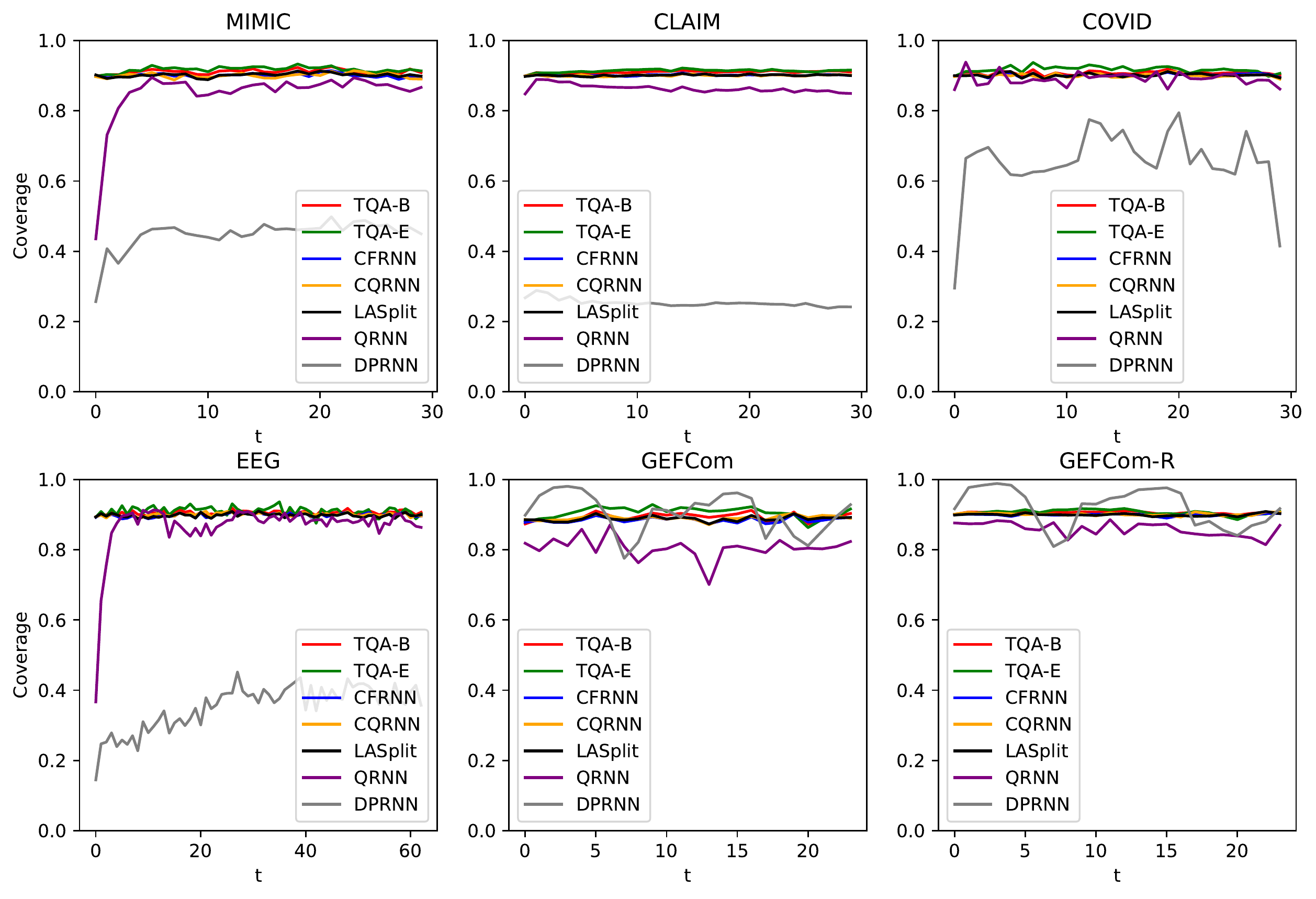}
\caption
{
Mean coverage rate at different $t$.
Conformal methods are valid at all $t$.
Coverage rate for non-conformal methods vary greatly through time.
\label{fig:appendix:cov_by_t}
}
\end{figure}
}
\FigAppendixCovByt

\textbf{Linear Regression}:
We repeat the experiments in Section~\ref{sec:exp} but replace the underlying LSTM with Linear Regression (by training one model for each $t$, taking input from up to $t-1$).
The \dataCLAIM data is skipped in this case due to difficulty in training.
The results are in Table~\ref{table:exp:appendix:linear_regression} and the conclusion stays the same.

\def\TabAppendixLinearRegression{

\begin{table}[ht]
\captionof{table}{
Mean coverage, tail coverage, and inverse efficiency with linear regression being the base point estimator.
Valid mean coverage and the best of tail coverage and inverse efficiency are in \textbf{bold}.
The conclusion is the same as the case of LSTM.
}
\centering
\begin{scriptsize}
\setlength\tabcolsep{6pt}
\begin{tabular}{l|ll|lll}
\toprule
Coverage     &\methodBudget & \methodError & \baselineCFRNN & CQR & \baselineMADSplit  \\
\midrule
\dataMIMIC & \textbf{90.65$\pm$1.36} & \textbf{91.63$\pm$0.79} & \textbf{89.75$\pm$1.61} & \textbf{89.96$\pm$1.62} & \textbf{89.81$\pm$1.48}\\
\dataCOVID & \textbf{90.93$\pm$1.45} & \textbf{91.95$\pm$0.86} & \textbf{90.28$\pm$1.71} & \textbf{90.17$\pm$1.57} & \textbf{90.25$\pm$1.55}\\
\dataEEG & \textbf{90.76$\pm$1.34} & \textbf{90.78$\pm$0.84} & \textbf{89.92$\pm$1.56} & \textbf{90.26$\pm$1.93} & \textbf{90.14$\pm$1.32}\\
\dataGEFCom & 89.66$\pm$0.17 & \textbf{90.46$\pm$0.12} & 89.32$\pm$0.14 & 88.98$\pm$0.28 & 89.21$\pm$0.16\\
\dataGEFComR & \textbf{90.57$\pm$0.63} & \textbf{90.78$\pm$0.42} & \textbf{90.08$\pm$0.72} & \textbf{90.13$\pm$0.66} & \textbf{90.12$\pm$0.59}\\
\midrule
Tail Coverage Rate $\uparrow$    &\\
\midrule
\dataMIMIC & 68.80$\pm$3.88 & \textbf{80.73$\pm$1.52} & 60.59$\pm$5.64 & 59.40$\pm$5.23 & 64.36$\pm$4.88\\
\dataCOVID & 70.38$\pm$4.04 & \textbf{82.20$\pm$1.51} & 64.35$\pm$6.00 & 68.30$\pm$4.75 & 69.41$\pm$4.94\\ 
\dataEEG & 68.11$\pm$2.73 & \textbf{77.66$\pm$1.27} & 61.09$\pm$3.96 & 54.45$\pm$6.58 & 66.81$\pm$2.70\\
\dataGEFCom & 73.72$\pm$0.92 & \textbf{81.58$\pm$0.33} & 69.31$\pm$0.95 & 70.01$\pm$1.12 & 71.40$\pm$0.92\\
\dataGEFComR & 75.34$\pm$0.88 & \textbf{81.61$\pm$0.81} & 70.69$\pm$1.42 & 72.69$\pm$1.39 & 73.01$\pm$1.48\\
\midrule
Inverse Efficiency $\downarrow$    &\\
\midrule
\dataMIMIC & 3.235$\pm$0.263 & 4.092$\pm$0.768 & 3.205$\pm$0.266 & \textbf{2.899$\pm$0.252} & 3.326$\pm$0.286\\
\dataCOVID & \textbf{0.898$\pm$0.028} & 1.129$\pm$0.171 & \textbf{0.893$\pm$0.030} & 0.921$\pm$0.042 & \textbf{0.894$\pm$0.028}\\
\dataEEG & \textbf{1.840$\pm$0.048} & 2.232$\pm$0.152 & \textbf{1.832$\pm$0.057} & 1.976$\pm$0.080 & 1.995$\pm$0.059\\
\dataGEFCom & 0.309$\pm$0.005 & 0.350$\pm$0.007 & \textbf{0.307$\pm$0.005} & 0.373$\pm$0.011 & 0.334$\pm$0.006\\
\dataGEFComR & \textbf{0.297$\pm$0.006} & 0.318$\pm$0.009 & \textbf{0.296$\pm$0.007} & 0.355$\pm$0.008 & 0.315$\pm$0.006\\
\bottomrule
\end{tabular}
\label{table:exp:appendix:linear_regression}
\end{scriptsize}
\end{table}
}
\TabAppendixLinearRegression

\textbf{Examples of Actual PIs}
Figure~\ref{fig:appendix:actual_pi_ts} how \methodBudget and \methodError improve the coverage for the least-covered TSs over \baselineCFRNN (as they use the same nonconformity score).
The rest of the baselines are in Figure~\ref{fig:appendix:actual_pi_ts2}.
We use random seed 1 and pick the 1-th-least-covered TS (according to the mean coverage across all baselines) from each dataset.
In general, \methodname would increase the width (by lowering $a_t$) if the first few observations of a TS shows some extremity, which is probably most noticeable in \dataCLAIM.
\baselineCQR and \baselineMADSplit also adapt to the widths by using different nonconformity scores.
Again, we note that the quality of \methodBudget might be further improved with better prediction $\hat{r}$.
\def \FigAppendixActualPITS{
\begin{figure}[ht]
    \centering
    \includegraphics[width=1\textwidth]{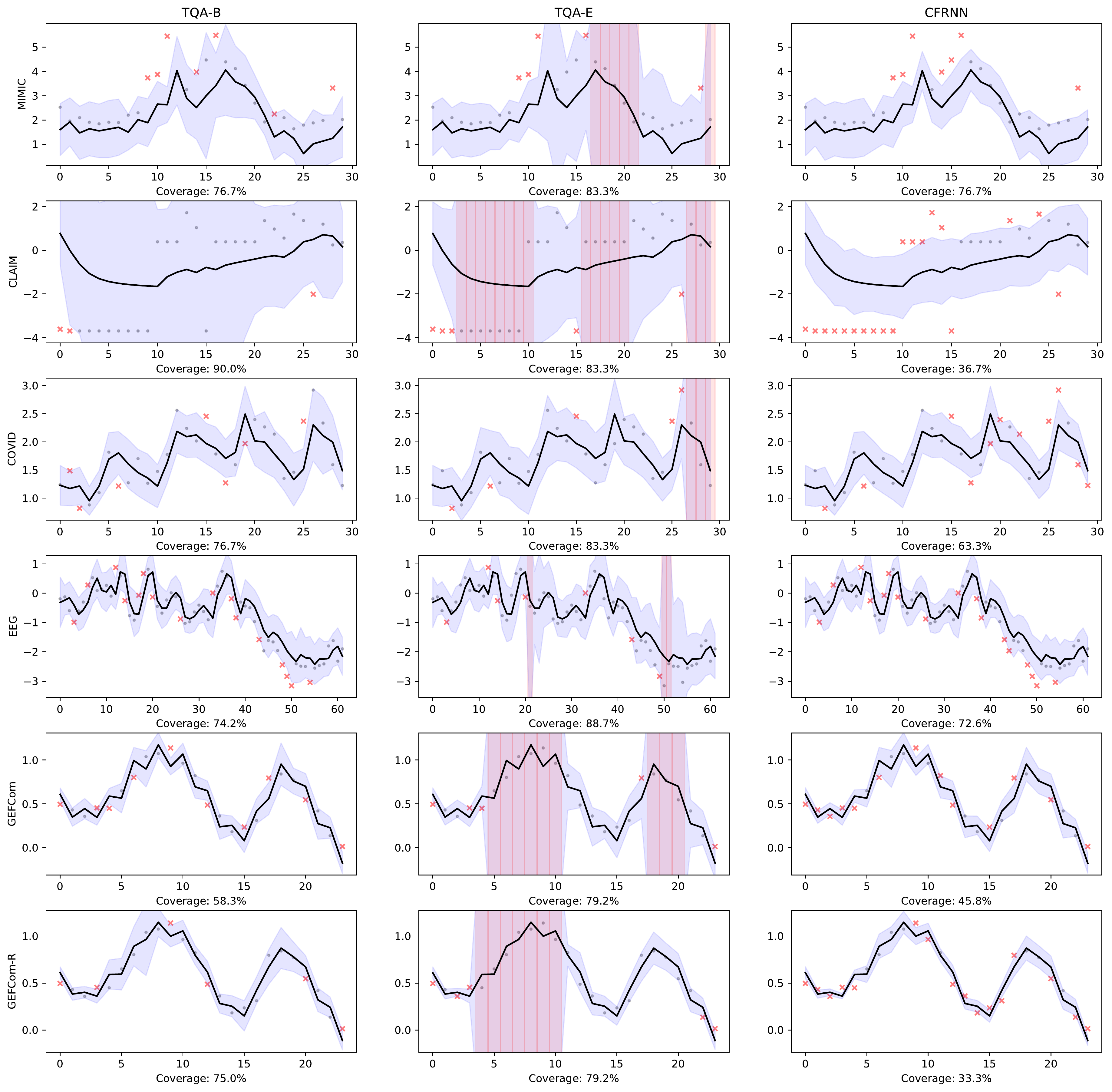}
\caption
{
Dark lines are $\hat{y}$, \colorbox{blue!15}{blue} bands are the PIs, and \fontred{\texttimes} denotes $y$ outside the PI.
\colorbox{red!15}{red} bands means infinitely-wide PIs for \methodError.
The adaptive-ness of \methodBudget and \methodError may be most obvious in \dataCLAIM and \dataGEFComR. 
We also note that \methodError seems to produce quite a few infinitely-wide PIs on these least-covered TSs (somewhat as expected). 
\label{fig:appendix:actual_pi_ts}
}
\end{figure}
}
\FigAppendixActualPITS
\def \FigAppendixActualPITSb{
\begin{figure}[ht]
    \centering
    \includegraphics[width=1\textwidth]{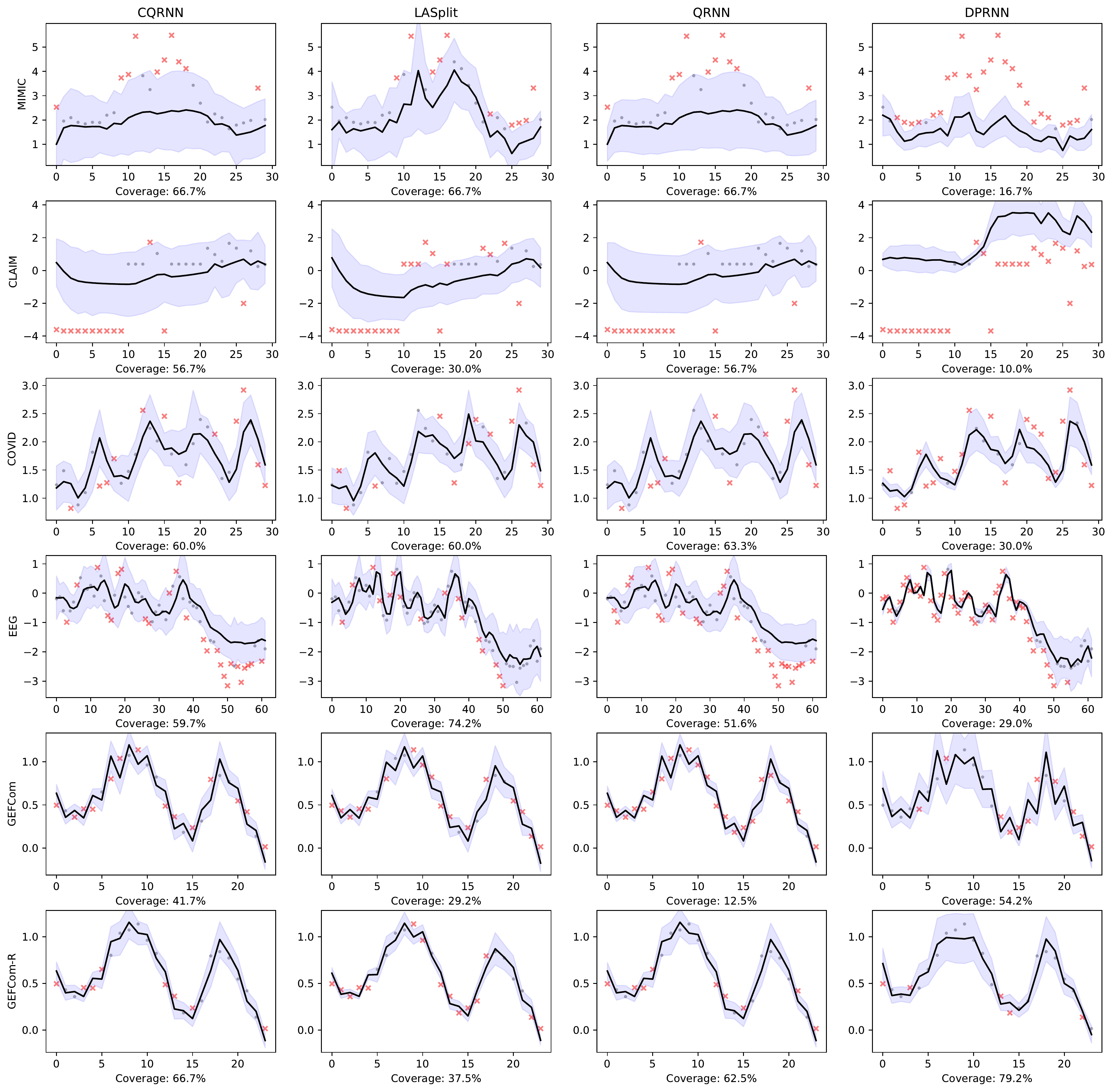}
\caption
{
Like Figure~\ref{fig:appendix:actual_pi_ts}, but for different baselines.
\label{fig:appendix:actual_pi_ts2}
}
\end{figure}
}
\FigAppendixActualPITSb

\end{document}